\documentclass[10pt,twocolumn,letterpaper]{article}
\pdfoutput=1
\usepackage[accsupp]{axessibility} 
\usepackage[pagenumbers]{cvpr}

\usepackage[dvipsnames]{xcolor}

\definecolor{citecolor}{HTML}{0071bc}
\definecolor{frontcolor}{HTML}{325ea5}
\definecolor{sidecolor}{HTML}{a58b77}
\definecolor{DeltaColor}{rgb}{0.039,0.73,0.71}
\definecolor{SigmaColor}{rgb}{0.98,0.45,0.0}
\definecolor{AlphaColor}{rgb}{0,0,0.8}
\definecolor{BetaColor}{rgb}{0.8,0,0.8}
\definecolor{GammaColor}{rgb}{0.514,0.34,0.224}
\definecolor{EpsilonColor}{rgb}{0.353,0.725,0.906}
\definecolor{PurpleColor}{HTML}{bca5ea}
\definecolor{OrangeColor}{rgb}{0.914,0.541,0.0.141}
\definecolor{GreenColor}{rgb}{0.137,0.573,0.565}
\definecolor{RedColor}{rgb}{0.949,0.275, 0.224}
\definecolor{LightCyan}{rgb}{0.88,1,1}
\definecolor{Gray}{gray}{0.85}

\def\mF{{\mathcal F}}

\def\mH{{\mathcal H}}

\def\mO{{\mathcal O}}

\def\mV{{\mathcal V}}

\def\0{{\mathbf  0}}
\def\1{{\mathbf  1}}

\def\bp{{\mathbf  p}}

\def\bp{{\mathbf  p}}

\newcommand{\highlight}[1]{{\color{black} #1}}

\newcommand{\xmark}{\textcolor{RedColor}{\ding{55}}\xspace}
\newcommand{\cmark}{\textcolor{GreenColor}{\ding{51}}\xspace}

\newlength\savewidth

\newcommand{\twoD}{2D\xspace}
\newcommand{\threeD}{3D\xspace}

\newcommand{\bodyMesh}{\mathcal{M}}

\newcommand{\cloNormFeat}{\mathcal{F}_\text{n}^\text{c}}
\newcommand{\bodyNormFeat}{\mathcal{F}_\text{n}^\text{b}}

\newcommand{\body}{\text{b}}

\newcommand{\sexynamefullname}{clothed human reconstruction with high- and low-frequency information}
\newcommand{\sexyname}{HiLo}

\def\eg{\emph{e.g.,}} 
\def\ie{\emph{i.e.,}} 
\def\cf{\emph{c.f.}} 
 
\def\wrt{{w.r.t.}} 

\newcommand{\oursdf}{\mathcal{H}(s;\beta)}
\newcommand{\ourmlp}{\phi_{si}}
\newcommand{\ourvoxel}{\mathcal{M}_v^{3D}}

\usepackage{pifont}
\usepackage[dvipsnames]{xcolor}
\usepackage{lipsum}
\usepackage{color, colortbl}
\usepackage{amsmath}
\usepackage{amsthm}
\usepackage{wrapfig}
\usepackage{array}
\usepackage{newlfont}
\usepackage{textcomp}
\usepackage{multirow}
\usepackage{commath}
\usepackage{hhline}
\usepackage{tabularx}
\usepackage{bigstrut}
\usepackage[para,online,flushleft]{threeparttable}
\usepackage{etoc}
\usepackage[ruled,linesnumbered]{algorithm2e}

\usepackage[toc,page,header]{appendix}
\usepackage{minitoc}

\definecolor{cvprblue}{rgb}{0.21,0.49,0.74}
\usepackage[pagebackref,breaklinks,colorlinks,citecolor=cvprblue]{hyperref}

\def\paperID{13536} 
\def\confName{CVPR}
\def\confYear{2024}
\title{HiLo: Detailed and Robust 3D Clothed Human Reconstruction with \\
High-and Low-Frequency Information of Parametric Models}

\author{Yifan Yang$^{1,2}$\footnotemark[1]~~Dong Liu$^{1}$\footnotemark[1]~~Shuhai Zhang$^{1,2}$~~Zeshuai Deng$^1$~~Zixiong Huang$^1$~~~Mingkui Tan$^{1,2,3}$\footnotemark[2]\\
$^1$South China University of Technology~~$^2$Pazhou Lab\\
$^3$Key Laboratory of Big Data and Intelligent Robot, Ministry of Education\\
{\tt\small \{seyoungyif, sesmildong, mszhangshuhai, sedengzeshuai, sesmilhzx\}@mail.scut.edu.cn} \\ {\tt\small mingkuitan@scut.edu.cn}
}
\begin{document}
\doparttoc %
\faketableofcontents %
\maketitle
\renewcommand{\thefootnote}{\fnsymbol{footnote}}
\footnotetext[2]{Corresponding author, *Equal contribution.} 
\renewcommand{\thefootnote}{\arabic{footnote}}
\begin{abstract}
Reconstructing 3D clothed human involves creating a detailed geometry of individuals in clothing, with applications ranging from virtual try-on, movies, to games.
To enable practical and widespread applications, recent advances propose to generate a clothed human from an RGB image. However, they struggle to reconstruct detailed and robust avatars simultaneously. 
We empirically find that the high-frequency (HF) and low-frequency (LF) information from a parametric model has the potential to enhance geometry details and improve robustness to noise, respectively.  Based on this, we propose \sexyname, namely \sexynamefullname, which contains two components. 1) To recover detailed geometry using HF information, we propose a progressive HF Signed Distance Function to enhance the detailed 3D geometry of a clothed human. We analyze that our progressive learning manner alleviates large gradients that hinder model convergence. 2) To achieve robust reconstruction against inaccurate estimation of the parametric model by using LF information, we propose a spatial interaction implicit function. This function effectively exploits the complementary spatial information from a low-resolution voxel grid of the parametric model.  
Experimental results demonstrate that \sexyname~outperforms the state-of-the-art methods by $10.43\%$ and $9.54\%$ in terms of Chamfer distance on the Thuman2.0 and CAPE datasets, respectively. Additionally, \sexyname~demonstrates robustness to noise from the parametric model, challenging poses, and various clothing styles. \footnote{Code link: \url{https://github.com/YifYang993/HiLo.git}}
\end{abstract}
    
\section{Introduction}
\label{sec:intro}
\begin{figure}[t]
    \centering
    \includegraphics[width=0.45\textwidth]{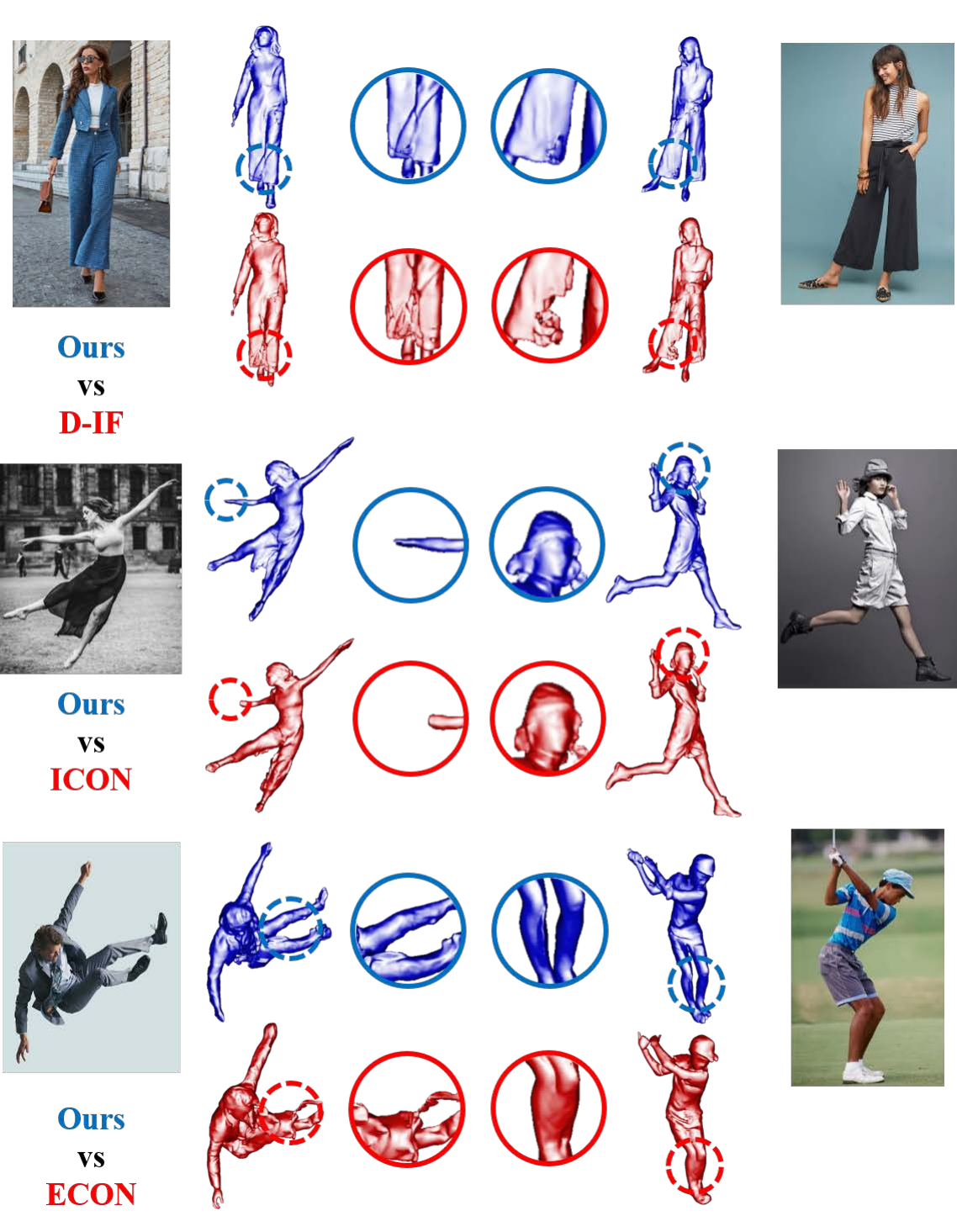}
    \caption{Visualization comparisons on in-the-wild images, our \sexyname~achieves more accurate and detailed reconstruction on challenging poses and diverse clothes.}
    \label{fig:vis_comp}
\end{figure}
The creation of 3D realistic digital human plays a pivotal role in the realm of mixed reality~\cite{jo2016effects, yoon2019effect, noh2015hmd}, remote presentation~\cite{yoon2019effect, aseeri2021influence}, film~\cite{sun2022research,galvane2015continuity}, and gaming~\cite{willumsen2018my}. Traditional methods often require expensive and specialized equipment combined with complex artistic efforts to customize the avatars\cite{schonberger2016pixelwise, guo2019relightables, zheng2014patchmatch}, which limits the ability of individuals to create personalized avatars easily. To address the limitation, recent approaches~\cite{alldieck2019learning, alldieck2019tex2shape, CAPEma2020learning, saito2019pifu, saito2020pifuhd, moon20223d, li2022dig, zheng2021pamir, xiu2022icon, xiu2023econ, yang2023dif} capture a 3D avatar from an RGB image of a clothed human, thus eliminating the need for costly scanning equipment and making it easier for a broader range of users to create personalized avatars. 

Despite the convenience of recent advances, the input image usually lacks details about delicate human body parts and diverse clothes from multiple angles. Moreover, the limited viewpoints and lack of accurate depth information make the reconstruction vulnerable to noise, \eg~inaccurate shape and pose of the estimated parametric body model~\cite{zhang2021pymaf,zhang2023pymafx, eksombatchai2018pixie}. Therefore, a detailed and robust 3D human reconstruction from an RGB image is challenging.

In spite of the impressive results of the previous methods, they have not fully addressed the problem of \textit{detailed} and \textit{robust} reconstruction simultaneously. Specifically, PIFu~\cite{saito2019pifu} produces overly smoothed or non-human body shapes on the unseen side of the human from the input image. ECON~\cite{xiu2023econ} requires Poisson Surface Reconstruction and replacement of body parts, introducing an extent of computational overhead (\cf~Sec.~\ref{sec:discuss}). Additionally, there is a risk of body part misalignment when the mid-term data is inaccurate. Considering that clothes need to conform to the surface of naked bodies, the geometry of the parametric model provides effective semantic regularization for reconstructing clothed humans. PaMIR~\cite{zheng2021pamir}, ICON~\cite{xiu2022icon}, and D-IF~\cite{yang2023dif} use parametric human bodies ~\cite{loper2015smpl, pavlakos2019smplx} to regularize the reconstruction. However, the performance of these methods degrades significantly when facing noise on the parameters from the estimated naked bodies (\cf~Sec.~\ref{sec:discuss}). 
\begin{figure}[t]
      \centering
    \includegraphics[width=0.45\textwidth]{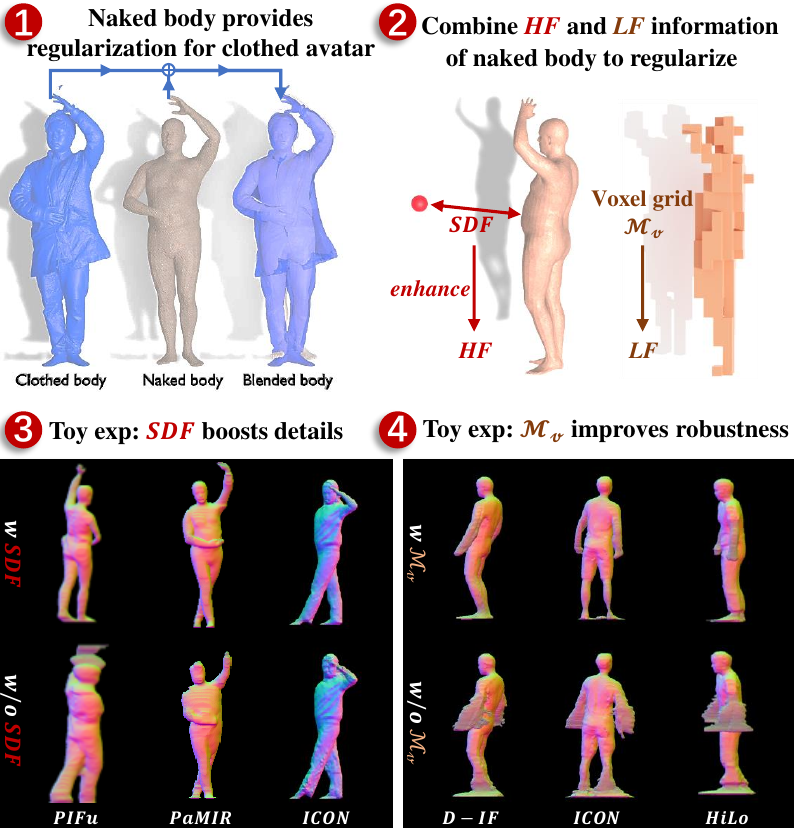}
    \caption{We empirically demonstrate the effectiveness of the high-frequency (HF) regularization from naked bodies in enhancing geometry details in Toy Experiment \includegraphics[width=0.8em]{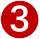}. We also verify the effectiveness of the low-frequency (LF) regularization in improving robustness to noise in Toy Experiment \includegraphics[width=0.8em]{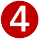}.}
    
    \label{fig:motivation}  
\end{figure}

To achieve \textit{robust} 3D clothed human reconstruction with \textit{detailed} geometry, we aim to explore how to further use the regularization from the parametric model to facilitate this goal.
Our exploration is based on two common observations.
First, \textit{high-frequency (HF) information} enhances details~\cite{rahaman2019spectral, mildenhall2020nerf}.
Considering that Signed Distance Function (SDF)~\cite{park2019deepsdf} describes the geometry of a parametric model by representing a distance to the object surface boundary, we investigate the effectiveness of SDF in improving the geometry details of clothed humans.
Second, \textit{low-frequency (LF) information} is relatively robust to noise~\cite{bu2023towards, chen2022rethinking, li2020wavelet, zhang2019interpreting}. 
Since inaccurate parametric model estimation within an error range has an insignificant impact on the corresponding low-resolution voxel grid~\cite{sun2022voxelgrid}, we seek to use the voxel grid to mitigate the noise of the estimated body.
As shown in Fig.~\ref{fig:motivation}, qualitative results demonstrate that SDF boosts the reconstruction details while the voxel grid improves robustness to noise.
However, how to effectively combine high- and low-frequency information to generate details and mitigate noise simultaneously is still an open question.

In this paper, we propose a high- and low-frequency paradigm \emph{\sexyname}, which stands for \emph{\sexynamefullname}. 
To achieve HF detail, we further enhance SDF with HF function~\cite{rahaman2019spectral}. Intuitively, by amplifying the variation of adjacent points that share similar SDFs, we allow for better delineation and capturing of fine details in the 3D human. Moreover, to alleviate the convergence difficulty caused by the large gradients amplified by HF function (\cf~Sec.~\ref{sec:pghfsdf}), we introduce a progressive HF SDF that learns detailed 3D geometry in a coarse-to-fine manner. To achieve robustness, we seek to capture the LF complementary information of the low-resolution voxel grid from the naked human body. To this end, we design a spatial interaction implicit function, which promotes the interaction of global and local information across different voxels via an attention mechanism.

We qualitatively and quantitatively evaluate our \sexyname~on in-the-wild images and benchmark datasets.
The experimental results verify the superiority of \sexyname~over previous approaches in three key aspects: 1) 3D geometry details (see Fig.~\ref{fig:vis_comp}). %
2) Robust reconstruction. %
3) Convergence speed. %
We summarize our contributions in three folds:
\begin{itemize}
    \item To enhance the geometry details and improve robustness against noise during the clothed human reconstruction process, we propose to explore the high-frequency (HF) information and low-frequency (LF) information from a parametric body model simultaneously.
    \item To facilitate detailed reconstruction, we introduce a progressive HF function to enhance the signed distance function (SDF) of a parametric model, providing regularization during the reconstruction process. This function learns an HF SDF in a progressive manner to alleviate the convergence difficulty associated with HF information. Experimental results show that \sexyname~reconstructs a more detailed clothed human.
    \item To ensure robust reconstruction, we employ LF information of low-resolution voxel grids from the parametric model to regularize the reconstruction. We propose a spatial interaction implicit function that reasons complementary information between different voxels. Experimental results show that \sexyname~is robust to various levels of noise.
    
\end{itemize}

\section{Preliminaries}
\label{SEC:PREL}
\subsection{Signed Distance Function} 
     Signed distance function (SDF)~\cite{park2019deepsdf} is a continuous function that takes a given spatial point $p$ with spatial coordinate $x \in \mathbb{R}^{n}$ and outputs the distance $s \in \mathbb{R}$ of the point to the closest point on the surface $\partial \Omega$ of an object $\Omega$:
\begin{equation} \label{eqn:sdf}
\begin{aligned}
   \mathcal{F}_{s}(\mathbf{p}) = s,~~
   s=\left\{ 
    \begin{array}{rc}
        d(x, \partial \Omega) & \mathrm{if}~x \not\in \Omega, \\
        -d(x, \partial \Omega) & \mathrm{if}~x \in \Omega, \\
    \end{array}
\right. \\
\end{aligned}
\end{equation}
where $d(x, \partial \Omega)=\mathop{\inf}\limits_{y \in \partial \Omega}(|| x-y||_2)$.  The sign of the distance implies whether the point is inside (negative) or outside (positive) of the surface $\partial \Omega$. $s=0$ denotes the point $\mathbf{p}$ locates on the $\partial \Omega$.

\subsection{Voxel Grid and Mesh Voxelization}
\textbf{Voxel Grids} %
$\Omega_v \in \mathbb{R}^{d \times h \times w}$ is a representative data structure for describing a 3D object $\Omega$. Specifically, $\Omega_v$ is a three-dimensional matrix of 3D space with depth $d$, height $h$ and width $w$.  $\Omega_v$ is composed of equally distributed and equally sized cube-shaped volumetric elements called voxels. The term voxel is the 3D counterpart to a 2D pixel. The resolution of a voxel grid is determined by the size of the voxels and the dimensions of the grid. Lower resolution implies larger voxels, resulting in a coarser representation of the space. 
Thus, we use the low-resolution voxel grid to represent the low-frequency information of a 3D object.

\textbf{Mesh voxelization} $\mathcal{V}$ is a computational technique that plays a crucial role in converting irregular continuous 3D geometric models~\cite{anguelov2005scape, loper2015smpl, pavlakos2019smplx} such as triangular mesh and point clouds into regular and discrete voxel grids. In this process, a 3D mesh $\Omega_m$, which is a collection of connected triangles, is converted into a grid of voxels $\Omega_v \in \mathbb{R}^{3}$.
In this paper, we use $\mathcal{V}$ to transfer SMPL-X mesh $\mathcal{M}$ to a low-resolution voxelized mesh $\mathcal{M}_{v} \in \mathbb{R}^{32 \times 32 \times 32}$, and then use a 3D CNN to encode $\mathcal{M}_{v}$ for $\mathcal{M}_{v}^{3D} \in $ for more flexibility. Based on our observation from Tab.~\ref{tab:smplnoise}, the low-frequency natural of $\mathcal{M}_{v}^{3D}$ aids multiple existing methods in mitigating various noise levels in SMPL-X shape and pose.

\section{Clothed Human Reconstruction with High- and Low-frequency Information} \label{sec:pghfsdf}

\begin{figure*}[th]
      \centering
    \includegraphics[width=1\textwidth]{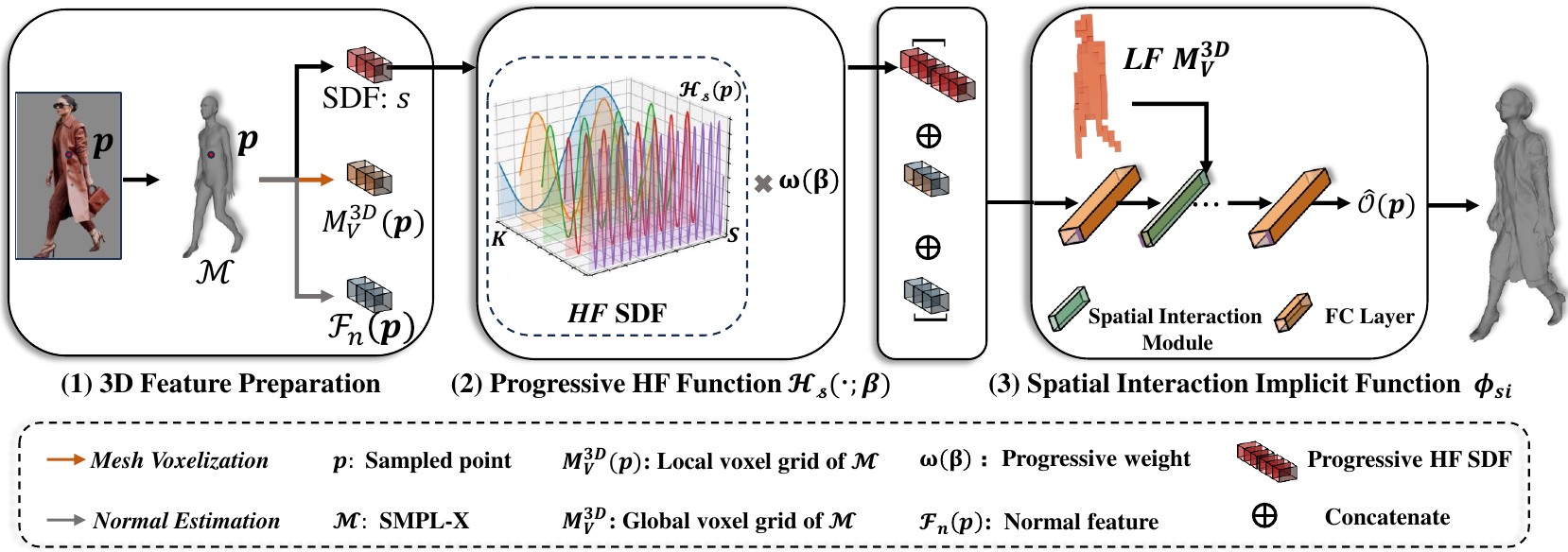}
    \caption{
    Overview of our proposed \sexyname. 
    Conditioned on a single-view image $\mathcal{I}$ and the corresponding SMPL-X $\mathcal{M}$, we first prepare a signed distance field $s$ and a low-resolution voxel grid $\mathcal{M}_v^{3D}$ of the naked body. Then, our proposed progressive high-frequency signed distance function $\mathcal{H}(s;\beta)$ enhances $s$ for detailed geometry of the clothed human and alleviates convergence difficulties introduced by large gradients in a coarse-to-fine learning manner. Moreover, we design an implicit function $\phi_{si}$ which leverages the complementary information of low-frequency voxels from $\mathcal{M}_v^{3D}$ to mitigate various levels of noise. Finally, we combine the above HF and LF features to $\phi_{si}$ to infer the occupancy field $\mathcal{\hat{O}}$ of the clothed avatar.}
    \label{fig:pipeline}  
\end{figure*}

We aim to robustly infer detailed 3D clothed avatars from RGB images $\mathcal{I}$. 
Recent advances~\cite{bhatnagar2020combining, bhatnagar2020loopreg, huang2020arch, he2021arch++} tend to use parametric naked body $\mathcal{M}$ such as SMPL~\cite{loper2015smpl} or SMPL-X~\cite{pavlakos2019smplx} estimated from $\mathcal{I}$ to provide semantic regularization on clothed human avatars. We empirically verify that high-frequency (HF) and low-frequency (LF) information from $\mathcal{M}$ are able to refine geometry and improve robustness to the reconstruction of the clothed human (Sec.~\ref{SEC:TOYEXP}). Based on this, we propose \sexynamefullname, namely \sexyname, which balances the HF and LF information to achieve detailed and robust reconstruction simultaneously.
As shown in Fig.~\ref{fig:pipeline}, \sexyname ~contains two key components: (1) To refine the geometry of clothed human with HF information, we propose to use \emph{progressive high-frequency function} to enhance the signed distance function (SDF) of $\mathcal{M}$ (\cf~Sec. \ref{SEC:PREL}), and alleviate the convergence difficulty introduced by large gradients in a coarse-to-fine enhancement manner. (2) To achieve robust reconstruction using low-frequency information, we explore complementary information from low-resolution voxels $\mathcal{M}_v$ from $\mathcal{M}$ for a more comprehensive understanding of human geometry.  To this end, we design a \emph{spatial interactive implicit function} (\cf~Sec. \ref{sec:simlp}) that leverages spatial information from local and global voxelized SMPL-X to predict the occupancy field $\mathcal{\hat{O}}$. Finally, we use the Marching Cubes (\cite{lorensen1998marching}) to obtain the 3D mesh of the clothed avatar from $\mathcal{\hat{O}}$. 

The overall optimization of our proposed \sexyname~minimizes the following objective function: 
\begin{equation} \label{eqn:loss}
    \mathcal{L}_{overall} = \mathcal{L}_{a}(\mathcal{\hat{O}}, \mathcal{O}), 
\end{equation}
where $\mathcal{L}_{a}$ is the MSE loss and $\mathcal{O}$ is a GT occupancy field.

\subsection{Progressive Growing High-Frequency SDF} \label{secSIIF}
Given that SDFs can enhance 3D reconstruction quality as confirmed in Sec.~\ref{SEC:TOYEXP}, we will leverage this for more realistic avatar reconstruction. However, directly fitting input coordinates with SDFs may lead to subpar representation of \textit{high-frequency} variation in geometry (see Sec.~\ref{sec:ablation}).  This aligns with previous work \cite{rahaman2019bias} indicating neural networks prioritize learning \textit{low-frequency} signals.  We will explore effective strategies to mitigate this.

\textbf{Conventional high-frequency SDF.} To 
improve the ability to represent complicated 3D shapes robustly, a straightforward way is to apply periodic functions $\mathcal{H}$ such as sine and cosine \cite{sitzmann2020siren, huang2024g} to extract high-frequency signals
on SDF of each sampled point via 
\begin{equation}
\begin{aligned}
\mathcal{H}(s)&=[s, \mathcal{H}_{0}(s), \mathcal{H}_{1}(s), \ldots,\mathcal{H}_{k}(s), \ldots,\mathcal{H}_{L}(s)], \\
\mathcal{H}_{k}(s)&=[\sin (2^{k} \pi s), \cos (2^{k} \pi s)], k\in\{0, 1,\ldots, L\}.\\
\end{aligned}
\end{equation}
In this way, we amplify the variation of adjacent points that share similar SDFs, allowing for better delineation and capturing of fine details in the 3D object.

Despite the positive characteristic of high-frequency SDF,  effective updating for parameters is difficult. Specifically, 
the gradient of $\mathcal{H}_k(s)$ \wrt~$s$ is calculated by
\begin{equation}
\label{eqn:jac_pos_enc}
  \frac{\partial \mathcal{H}_k(s)}{\partial s} = 2^{k}\pi [\cos{(2^{k}\pi s)}, -\sin{(2^{k}\pi s)}].
\end{equation}
Eqn. (\ref{eqn:jac_pos_enc}) incorporated with the coefficient $2^{k}\pi$ will significantly amplify the gradient signals regarding SDF, especially for larger $k$.
Large gradients could lead to convergence difficulties and numerical instability, ultimately resulting in poor representation performance \cite{pascanu2013difficulty,allen2019convergence}.

\begin{figure}[th]
    \centering
    \includegraphics[width=0.4\textwidth]{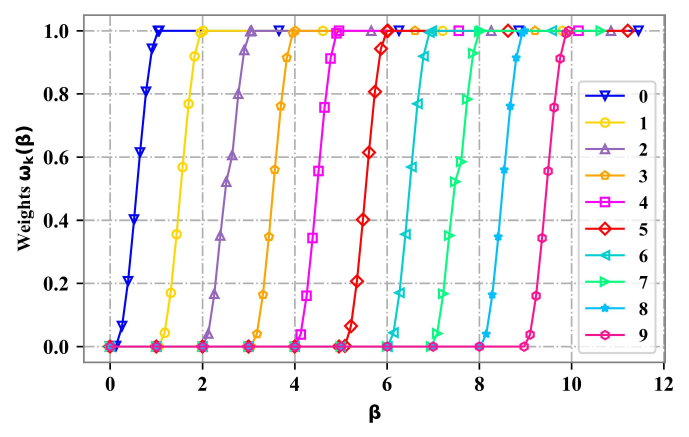}
    \caption{Illustration of the relationship between progressive weights $\omega$ and $\beta$ during the training process.}
    \label{fig:weights}
\end{figure}

\textbf{High-frequency SDF in a growing manner.} To address the above issue, we introduce a progressively growing approach as shown in Fig.~\ref{fig:pipeline} (2),
initially emphasizing low-frequency signal learning and gradually focusing on learning the high-frequency geometry. 
Specifically, in the early stage of training, we reduce the weight of high-frequency signals (\eg $\mathcal{H}_k(s)$) which have higher $k$  and progressively increase their importance during training. We formulate this schedule as $\mathcal{H}_k(s;\beta)$ with a weight $\omega_k(\beta)$ following \cite{park2021nerfies}:
\begin{equation}
\begin{aligned}
      &\mathcal{H}_k(s; \beta) = \omega_k(\beta) [\sin{(2^{k}\pi s)},\cos{(2^{k}\pi s)}], \\
& \omega_k(\beta)=\left\{ 
    \begin{array}{ll}
        0  &\mathrm{if}~\beta -k< 0; \\
        \frac{1-\cos((\beta-k)\pi)}{2} &\mathrm{if}~0\leq\beta-k < 1;\\
        1  &\mathrm{if}~\beta-k>1, \\
    \end{array}
\right. \\
\end{aligned}
\end{equation}
where $\beta$ is proportional to the iteration of the optimization process, see Fig.~\ref{fig:weights} for the relationship between $ \omega_k(\beta)$ and $\beta$. With $\omega_k(\beta)$, the gradient of $\mathcal{H}_k(s;\beta)$ becomes
\begin{equation}
    \frac{\partial \mathcal{H}_k(s; \beta)}{\partial s} = \omega_k(\beta) 2^{k}\pi [\cos{(2^{k}\pi s)},-\sin{(2^{k}\pi s)}].
\end{equation}
Then, during the beginning of the optimization, $\beta$ is set so small that only frequency components with a smaller value of $k$ will be assigned a non-zero weight, while the frequency components with a higher value of $k$ will be omitted. Throughout the optimization, the higher-frequency components are progressively activated. This manner allows \sexyname~to explore the low-frequency part and later focus on the fine-grained geometry of 3D humans.

\subsection{Low-Frequency Information for Robustness} \label{sec:simlp} 
Most recent methods ~\cite{xiu2023econ, xiu2022icon, yang2023dif, zheng2021pamir} are based on SMPL-X. However, SMPL-X estimation often faces \textbf{misalignment issues} with the corresponding image, especially when facing a challenging human pose. Thus, it is crucial to \textbf{achieve robust reconstruction} against misaligned SMPL-X. 
Our results (\cf~Fig.~\ref{fig:motivation}) show that
the low-frequency information, represented by low-resolution voxel grids $\bodyMesh_v$ of SMPL-X $\bodyMesh$, enhances robustness against noise. We leverage this insight to incorporate local and global information of $\bodyMesh_v$ for improved regularization in reconstruction.

\textbf{Local voxels for 3D feature preparation.}
Motivated by that point-wise local 3D features from $\bodyMesh_v$ are robust to out-of-distribution pose and shape~\cite{zheng2021pamir},
we voxelize the estimated SMPL-X $\mathcal{M}$ and query it by $\mathbf{p}$. Specifically, to obtain the voxelization features, we convert the corresponding SMPL-X $\bodyMesh$ into a low-resolution voxel grid $\bodyMesh_{\mathcal{V}}$ by mesh voxelization operation $\mathcal{V}$~\cite{zheng2021pamir} and encode it via a 3D CNN $f_\text{\threeD}$ for a 3D feature volume $\bodyMesh^{3D}_{\mathcal{V}}$.
To obtain point-wise 3D features, we use trilinear interpolation to sample $\ourvoxel$ based on coordinate $\mathbf{p}$ of sampled 3D points, resulting in $\bodyMesh^{3D}_{\mathcal{V}}(\mathbf{p})$. We empirically find that by combining $\bodyMesh^{3D}_{\mathcal{V}}(\mathbf{p})$, \sexyname~is robust to SMPL-X noise (\cf~Sec. \ref{sec:discuss}).   
As shown in Fig.~\ref{fig:pipeline} (1), in addition to $\mathcal{H}(s;\beta)$ and $ \bodyMesh^{3D}_{\mathcal{V}}$,  we follow ICON~\cite{xiu2022icon} to use a normal feature $\mathbf{F}_{\mathrm{n}}(\mathbf{p})$ to provide detailed texture information.
Then, we concatenate them into one final feature $\mathbf{F}_{c}^1$=$[\mathcal{H}(s;\beta), \bodyMesh^{3D}_{\mathcal{V}}({\mathbf{p}}), \mathbf{F}_{\mathrm{n}}(\mathbf{p})]$ and then fed $\mathbf{F}_{c}^1$ to our designed implicit function to reconstruct the clothed avatar.

\textbf{Global voxels for spatial interaction implicit function.} To reconstruct clothed avatars, a typical solution is to map 3D features $\mathbf{F}$ to a continuous occupancy field that represents the interior and exterior of a clothed human.
To this end, numerous literature~\cite{saito2019pifu, zheng2021pamir, xiu2022icon, yang2023dif} uses an implicit function parameterized by a memory-efficient multi-layer perceptron (MLP) $\mathcal{T}$ to map $\mathbf{F}$ into an occupancy field $\mathcal{\hat{O}}$.

\begin{figure}[t]
    \centering
    \includegraphics[width=0.4\textwidth]{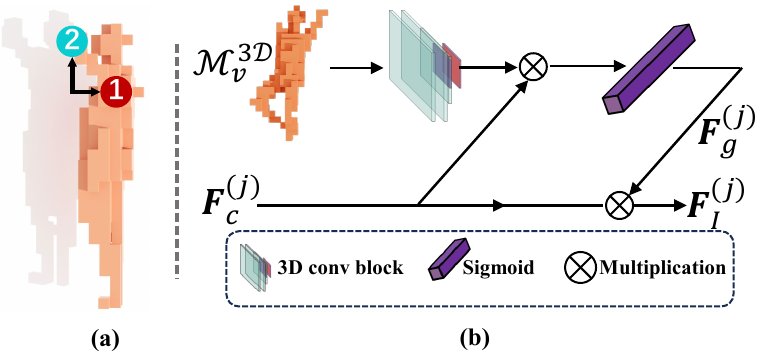}
    \vspace{-4pt}
    \caption{(a) Complementarity of voxel \includegraphics[width=0.8em]{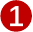} and \includegraphics[width=0.8em]{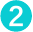}. (b) Illustration of the spatial interaction module $\mathcal{A}$.}
    \label{fig:simlp}
\end{figure}

However, the potential issue of the existing implicit function lies in its underutilization of the global information inherent in 3D data. Previous research \cite{bell2016inside,newell2016stacked,jaderberg2015spatial} has shown that the representation ability of features can be improved by capturing the global correlation between the features. For 3D clothed human reconstruction, different human body parts contain distinct yet complementary spatial information. For instance, as shown in Fig.~\ref{fig:simlp} (a), the voxels located on the shoulder (the red point) may offer valuable topological cues to constrain the prediction of geometry near the elbow (the blue point). 
\par To leverage global information from the voxel grid of SMPL-X of $\ourvoxel$, 
we design a spatial interaction module $\mathcal{A}$ into $\phi$ to infer the 3D occupancy, denoted by $\phi_{si}$, see detail in Appx \ref{sec_appx:detail_simlp}. 
As shown in Fig.~\ref{fig:simlp} (b), $\phi_{si}$ injects global-scale features of $\ourvoxel$ to the local 3D feature with the aim of introducing whole-body awareness to $\phi_{si}$: 
\begin{equation} \label{eqn:ourmlp}
    {\phi_{si}}(\mathbf{F}_{c}^1) \to \mathcal{\hat{O}},~~\phi_{si}(\cdot)=\mathcal{A}^{N+1} \circ T^{(N+1)} \circ  \cdots \circ \mathcal{A}^{1}(\cdot) \circ T^{(1)}.
\end{equation}

\textbf{Optimization.} %
We optimize parameters of $\ourmlp$ and $f_{3D}$ via minimizing the MSE loss in Eqn.~(\ref{eqn:loss}) between the predicted occupancy field $\mathcal{\hat{O}}$ and the ground-truth occupancy field $\mathcal{O}$.
With $\mathcal{\hat{O}}$, we reconstruct the triangular mesh of the 3D clothed avatar via marching cubes algorithm ~\cite{lorensen1998marching}.

\section{Experiments}
\noindent \textbf{Datasets:} 
We conduct experiments on two open-source datasets, \ie~Thuman2.0~\cite{zheng2019deephuman} and CAPE~\cite{CAPEma2020learning}, which both contain various human shapes with different human poses and diverse clothes. Specifically, following ICON~\cite{xiu2022icon}, the CAPE dataset is divided into the "CAPE-FP" and "CAPE-NFP" sets, which have "fashion" and "non-fashion" poses, respectively, to further analyze the generalization to complex body poses. 

Moreover, to evaluate our \sexyname~on in-the-wild images, we follow ICON to collect 200 diverse images from Pinterest\footnote{https://www.pinterest.com/}. The images contain humans performing dramatic movements or wearing diverse clothes.

\begin{figure*}
    \centering
    \includegraphics[width=0.9\linewidth]{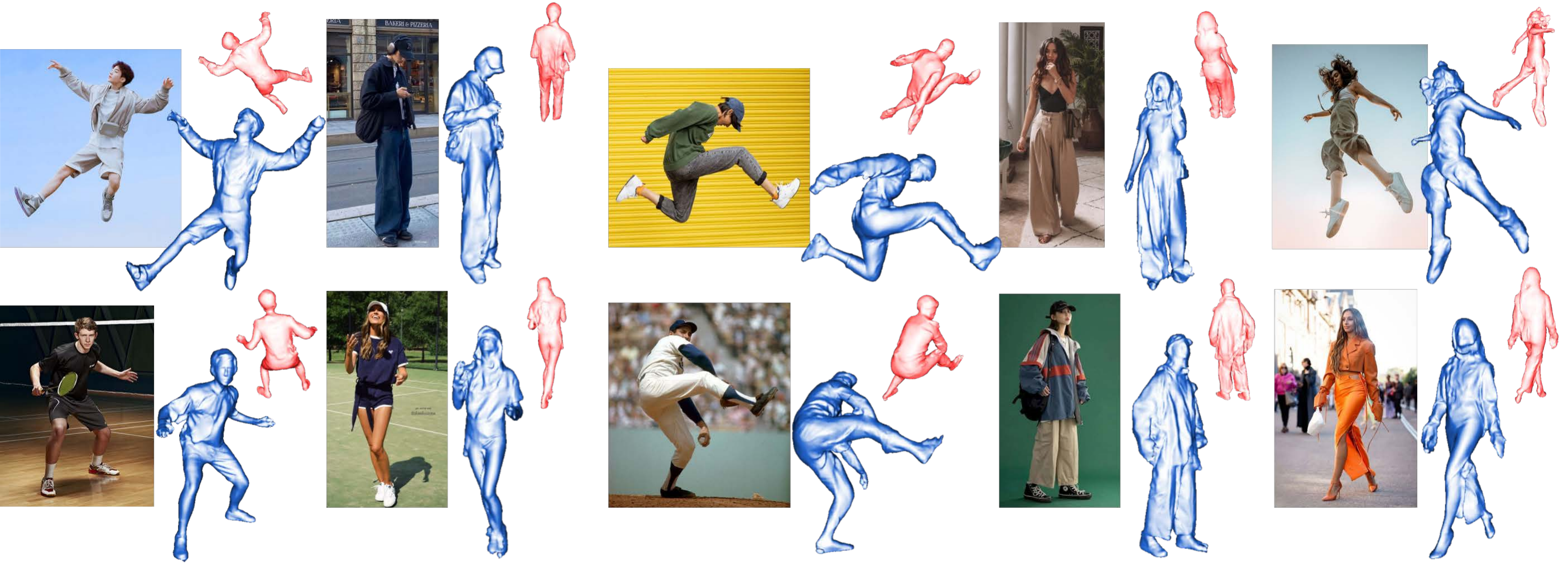}
    \caption{Visualization results of 3D clothed avatar reconstruction with our \sexyname~from in-the-wild images, which present various clothing and challenging poses. We show the front (\textcolor[RGB]{62,113,197}{blue}) and rotated (\textcolor[RGB]{195,46,46}{red}) views.}
    \label{fig:inthewild}
\end{figure*}

\begin{table*}[t]
  \centering
  \caption{Toy experiments about the impact of SDF on 3D clothed human reconstruction, on seen (\ie~training and test on the same dataset) and unseen (\ie~training on Thuman2.0 and test on CAPE) settings.}
     \begin{threeparttable}
 \small
\setlength{\tabcolsep}{3pt} 
\resizebox{0.9\linewidth}{!}{
    \begin{tabular}{c|ccc|ccc|ccc|ccc|ccc}
    \toprule
 Dataset & \multicolumn{3}{c}{CAPE-FP} & \multicolumn{3}{c}{CAPE-NFP} & \multicolumn{3}{c|}{CAPE} &\multicolumn{3}{c|}{Thuman2} & \multicolumn{3}{c}{CAPE} \\
\cmidrule{1-16} Methods & Chamfer ($\downarrow$) & P2S ($\downarrow$)  & Normals ($\downarrow$) & Chamfer & P2S & Normals & Chamfer & P2S & Normals & Chamfer & P2S & Normals & Chamfer & P2S  & Normals \\
    \midrule
           $\mathrm{PIFu}$~\cite{saito2019pifu} & 2.1000  & 2.0930  & 0.0910  & 2.9730  & 2.9400  & 0.1110  & 2.6820  & 2.6580  & 0.1040 & 2.6880  & 2.5730  & 0.0970  & 7.1244  & 2.7633  & 0.3902 \\ $\mathrm{PIFu}_{\mathrm{w}~\mathrm{SDF}}$ & 0.8908  & 0.8637  & 0.0676  & 0.9848  & 0.9545  & 0.0698  & 0.9437  & 0.9178  & 0.0707  & 1.6659  & 1.7934  & 0.1360  & 1.3078  & 1.4306  & 0.0980\\
\cmidrule{1-16} $\mathrm{PaMIR}$~\cite{zheng2021pamir} & 1.2250  & 1.2060  & 0.0550  & 1.4130  & 1.3210  & 0.0630  & 1.3500  & 1.2830  & 0.0600  & 1.4388 & 1.5613  & 0.1071  & 0.9339  & 0.9444  & 0.0659\\
           $\mathrm{PaMIR}_{\mathrm{w}~\mathrm{SDF}}$ & 0.9188  & 0.8788  & 0.0565  & 1.1132  & 1.0729  & 0.0611  & 1.0112  & 0.9725  & 0.0601 & 1.4073  & 1.5624  & 0.1174  & 0.8438  & 0.8179  & 0.0572 \\
          \cmidrule{1-16}
          ICON~\cite{xiu2022icon} & 0.7475  & 0.7488  & 0.0508  & 0.8656  & 0.8639  & 0.0545  & 0.8055  & 0.8084  & 0.0539 & 1.1431  & 1.3020  & 0.0923  & 0.8610  & 0.8878  & 0.0606 \\
    $\mathrm{ICON}_{\mathrm{w/o}~\mathrm{SDF}}$ & 1.0243  & 0.9478  & 0.0741  & 1.4862  & 1.3313  & 0.0919  & 1.2736  & 1.1538  & 0.0850  & 1.3114  & 1.2116  & 0.1015  & 7.4892  & 1.7708  & 0.3780 \\  %
    \bottomrule
    \end{tabular}%
    }
    \end{threeparttable}
  \label{tab:TOY}%
\end{table*}%

\begin{table*}[t]
  \centering
  \caption{(A) Comparison experiments and (B) ablation studies on seen (\ie~training and test on the same dataset) and unseen (\ie~training on Thuman2.0 and test on CAPE) settings. The \textbf{bold} and the \underline{underlined} numbers indicate the best and second-best results, respectively. "-" denotes that PIFuHD and ECON does not provide a training code.}
     \begin{threeparttable}
 \small
\setlength{\tabcolsep}{3pt} 
\resizebox{1\linewidth}{!}{
    \begin{tabular}{c|c|ccc|ccc|ccc|ccc|ccc}
    \toprule
    &\multicolumn{1}{c|}{} &\multicolumn{9}{c|}{Train on Thuman2.0 and test on CAPE}    & \multicolumn{6}{c}{Train and test on the same dataset (Thuman2.0 or CAPE)}\\
    \midrule
  \multirow{2}[4]{*}{Group}          & Dataset & \multicolumn{3}{c}{CAPE-FP} & \multicolumn{3}{c}{CAPE-NFP} & \multicolumn{3}{c|}{CAPE} &\multicolumn{3}{c|}{Thuman2.0} & \multicolumn{3}{c}{CAPE} \\
\cmidrule{2-17}          & Methods & Chamfer ($\downarrow$) & P2S ($\downarrow$)  & Normals ($\downarrow$) & Chamfer & P2S   & Normals & Chamfer & P2S  & Normals & Chamfer & P2S & Normals & Chamfer & P2S   & Normals \\
    \midrule
    \multirow{6}[2]{*}{A} %
          & PIFu~\cite{saito2019pifu} & 2.1000  & 2.0930  & 0.0910  & 2.9730  & 2.9400  & 0.1110  & 2.6820  & 2.6580  & 0.1040 & 2.6880  & 2.5730  & 0.0970  & 7.1244  & 2.7633  & 0.3902 \\
          &PIFuHD~\cite{saito2020pifuhd} &2.3020 &2.3350 &0.0900 &3.7040 &3.5170 &0.1230 &3.2370 &3.1230 &0.1120 &2.4613 &2.3605 &0.0924&-&-&-\\
          & PaMIR~\cite{zheng2021pamir} & 1.2250  & 1.2060  & 0.0550  & 1.4130  & 1.3210  & 0.0630  & 1.3500  & 1.2830  & 0.0600  & 1.4388 & 1.5613  & 0.1071  & 0.9339  & 0.9444  & 0.0659\\
          & ICON~\cite{xiu2022icon}  & \underline{0.7475}  & \underline{0.7488}  & 0.0508  & \underline{0.8656}  & \underline{0.8639}  & 0.0545  & \underline{0.8055}  & \underline{0.8084}  & 0.0539 & 1.1431  & 1.3020  & 0.0923  & 0.8610  & 0.8878  & \underline{0.0606} \\
          & ECON~\cite{xiu2023econ}  & 0.9651  & 0.9175  & \textbf{0.0412}  & 0.9983  & 0.9694  & \textbf{0.0415}  & 0.9872  & 0.9521  & \textbf{0.0414}  & -  & -  & -  & -     & -     & -\\
          & D-IF~\cite{yang2023dif}  & 0.8038  & 0.7766  & 0.0546  & 0.9877  & 0.9491  & 0.0611  & 0.8878  & 0.8574  & 0.0589 & \underline{1.0305}  & \underline{1.0864}  & \underline{0.0830}  & \underline{0.8332}  & \underline{0.8489}  & 0.0597 \\

\midrule
       \multirow{3}[1]{*}{B}   & $\mathrm{\sexyname}_{\mathrm{w/o}~ \mathcal{H}(s;\beta)}$ & 0.7564 & 0.7449 & 0.0514 & 0.8697  & 0.8658   & 0.0553  & 0.8118 & 0.8045 	& 0.0547 & 0.9442 & 1.0323 & 0.0785  & 0.7971  & 0.7999  & 0.0551\\
          & $\mathrm{\sexyname}_{\mathrm{w/o}~ \mathcal{\phi}_{si}}$ & 0.7996  & 0.7860  & 0.0569  & 0.9112  & 0.9042  & 0.0601  & 0.8555  & 0.8449  & 0.0468  & 0.9886  & 1.0836  & 0.0850  & 0.7999  & 0.7948  & 0.0547\\
          & $\mathrm{\sexyname}_{\mathrm{w/o}~ \mathcal{H}(s;\beta)~\mathrm{w/o}~ \mathcal{\phi}_{si}}$ & 0.8662  & 0.8970  & 0.0647  & 1.0201  & 1.0525  & 0.0706  & 0.9362  & 0.9720  & 0.0690 & 1.1220  & 1.2544  & 0.0954  & 0.8125  & 0.8224  & 0.0588 \\
\midrule
\multirow{1}[1]{*}{ }   & \cellcolor[rgb]{ .949,  .863,  .859}\sexyname~(Ours)  & \cellcolor[rgb]{ .949,  .863,  .859}\textbf{0.6954} & \cellcolor[rgb]{ .949,  .863,  .859}\textbf{0.6876} & \cellcolor[rgb]{ .949,  .863,  .859}\underline{0.0471} & \cellcolor[rgb]{ .949,  .863,  .859}\textbf{0.7830} & \cellcolor[rgb]{ .949,  .863,  .859}\textbf{0.7876} & \cellcolor[rgb]{ .949,  .863,  .859}\underline{0.0499} & \cellcolor[rgb]{ .949,  .863,  .859}\textbf{0.7430} & \cellcolor[rgb]{ .949,  .863,  .859}\textbf{0.7428} & \cellcolor[rgb]{ .949,  .863,  .859}\underline{0.0499} & \cellcolor[rgb]{ .949,  .863,  .859}\textbf{0.9230} & \cellcolor[rgb]{ .949,  .863,  .859}\textbf{0.9855} & \cellcolor[rgb]{ .949,  .863,  .859}\textbf{0.0732} & \cellcolor[rgb]{ .949,  .863,  .859}\textbf{0.7861} & \cellcolor[rgb]{ .949,  .863,  .859}\textbf{	0.7729} & \cellcolor[rgb]{ .949,  .863,  .859}\textbf{0.0544}\\
    \bottomrule
    \end{tabular}%
    }
    \end{threeparttable}
  \label{tab:comparison}%
\end{table*}%

\begin{table*}[htbp]
  \centering
  \caption{Toy experiments on 3D reconstruction with different levels of SMPL-X noise in terms of chamfer distance on unseen CAPE dataset. The voxel grid of naked body $\mathcal{M}_v^{3D}(p)$ improves the robustness of reconstruction.}
  \resizebox{0.9\linewidth}{!}{
    \begin{tabular}{l|c|ccc|ccc|ccc}
\cmidrule{1-11}    & & \multicolumn{3}{c}{SMPL-X Noise=0.1} & \multicolumn{3}{c}{SMPL-X Noise=0.2} & \multicolumn{3}{c}{SMPL-X Noise=0.5} \\
\cmidrule{1-11}   Methods & $\mathcal{M}_v^{3D}(p)$  & CAPE-FP & CAPE-NFP & CAPE & CAPE-FP & CAPE-NFP & CAPE  & CAPE-FP & CAPE-NFP & CAPE  \\
\cmidrule{1-11}    ICON~\cite{xiu2022icon} & \xmark      & 1.7949  & 2.0537  & 1.9284  & 2.6695  & 3.0917  & 2.9365  & 4.2181  & 4.8266  & 4.6716  \\
    $\mathrm{ICON}_{\mathrm{w}~\mathcal{M}_v^{3D}(p)}$ & \cmark  & 1.4381  & 1.5380  & 1.3411  & 2.1950  & 2.3067  & 2.0723  & 2.2129	& 2.3760	& 2.1188  \\
    \midrule 
$\mathrm{D\text{-}IF}$~\cite{yang2023dif} & \xmark    & 1.6078  & 1.7881  & 1.7037  & 2.6853  & 3.1002  & 2.9962  & 4.3591 & 4.8006	 &  4.7310  \\
    $\mathrm{D\text{-}IF}_{\mathrm{w}~\mathcal{M}_v^{3D}(p)}$ & \cmark  & 1.3557  & 1.7877  & 1.6064  & 1.8022  & 2.4110  & 2.1959  & 3.0307 & 4.4190 & 4.0807   \\
    \midrule
    $\mathrm{\sexyname}_{\mathrm{w/o}~\mathcal{M}_v^{3D}(p)}$ & \xmark  & 1.9518  & 2.1966  & 2.0877  & 2.9315  & 3.4561  & 3.2661  & 4.6382 & 	5.0606	 &  4.9844  \\
    $\mathrm{\sexyname}$ & \cmark & 1.0517  & 1.3210  & 1.2004  & 1.0893  & 1.5876  & 1.3427  & 1.0960  & 1.6156  & 1.3593  \\

    \bottomrule
    \end{tabular}}%
  \label{tab:smplnoise}%
\end{table*}%

\begin{table*}[htbp]
  \centering
  \caption{Ablations of our progressive high-frequency SDF on 3D reconstruction with different levels of SMPL-X noise in terms of chamfer distance on unseen CAPE dataset. In addition to $\mathcal{M}_v^{3D}$, our progressive high-frequency SDF is also to handle SMPL-X noise due to the coarse-to-fine learning manner.}
  \resizebox{1\linewidth}{!}{
    \begin{tabular}{l|cccc|ccc|ccc|ccc}
\cmidrule {1-14}   \multicolumn{2}{r}{} &  &  &  & \multicolumn{3}{c}{SMPL-X Noise=0.1} & \multicolumn{3}{c}{SMPL-X Noise=0.2} & \multicolumn{3}{c}{SMPL-X Noise=0.5} \\
\cmidrule{1-14}  Methods & $\mathcal{M}_v^{3D}(p)$      & $\mathcal{H}_s(p;\beta)$      & $\mathcal{H}_s(p)$ & $SDF$ & CAPE-FP & CAPE-NFP & CAPE & CAPE-FP & CAPE-NFP & CAPE  & CAPE-FP & CAPE-NFP & CAPE  \\

    $\mathrm{\sexyname}_{\mathrm{w/o}~\mathcal{H}_s(p;\beta)}$ & \cmark     & \xmark     & \xmark  & \cmark  & 1.1435  & 1.4700  & 1.3124  & 1.3401  & 1.9339  & 1.6909  & 1.2861  & 1.8200  & 1.5620  \\
    $\mathrm{\sexyname}_{\mathrm{w}~\mathcal{H}_s(p)}$ & \cmark     & \xmark     & \cmark   & \xmark & 1.1932  & 1.5541  & 1.3904  & 1.1243  & 1.5575  & 1.3701  & 1.2794  & 1.8973  & 1.6652  \\
        $\mathrm{\sexyname}$ & \cmark     & \cmark     & \xmark    &\xmark & 1.0517  & 1.3210  & 1.2004  & 1.0893  & 1.5876  & 1.3427  & 1.0960  & 1.6156  & 1.3593  \\

    \bottomrule
    \end{tabular}}%
  \label{tab:smplnoise_sdf}%
\end{table*}%

\noindent \textbf{Metrics:} 
We evaluate our \sexyname~and baseline methods in terms of three metrics: \textbf{Chamfer distance} and \textbf{P2S distance} mainly measure coarse geometry error, while \textbf{Normals} mainly captures high-frequency differences. See details in Appx.~\ref{sec_supp:metric}. 

\noindent \textbf{Baselines:} 
We compare our proposed \sexyname~with mainstream state-of-the-art methods, including PIFu~\cite{saito2019pifu}, PaMIR~\cite{zheng2021pamir}, ICON~\cite{xiu2022icon}, ECON~\cite{xiu2023econ} and D-IF~\cite{yang2023dif}, refer to the Appx.~\ref{sec_appx:baseline} for the detailed description. 
To demonstrate the necessity of naked 3D body regularization, we first conduct a toy experiment to study the effect of SDF on different baselines. 
To this end, we design three variants based on PIFu, PaMIR, and ICON, which incorporate SDF into existing methods (PIFu and PaMIR) or remove SDF from the existing method (ICON), namely 
$\mathrm{PIFu}_{\mathrm{w}~\mathrm{SDF}}$, 
$\mathrm{PaMIR}_{\mathrm{w}~\mathrm{SDF}}$ and
$\mathrm{ICON}_{{w/o}~\mathrm{SDF}}$, respectively.

For ablation studies, we construct several variants of our \sexyname: 
1) $\mathrm{\sexyname}_{{w/o}~\ourmlp}$ replaces our spatial interaction implicit function with the vanilla implicit function;
2) $\mathrm{\sexyname}_{{w/o}~\ourvoxel}$ is constructed by removing the voxelized SMPL-X;
3) $\mathrm{\sexyname}_{{w/o}~\oursdf}$ is constructed by replacing our progressive HF SDF with vanilla SDF.

\noindent \textbf{Implementation details:} 
We implement our approach using PyTorch\footnote{We will release our code.}~\cite{paszke2019pytorch} and train our networks with RMSprop~\cite{tieleman2012lecture} optimizer. For a fair comparison, we follow all common hyper-parameter settings same as ICON~\cite{xiu2022icon}. See more implementation details in the Appx.
\begin{figure}
    \centering
    \includegraphics[width=0.5\textwidth]{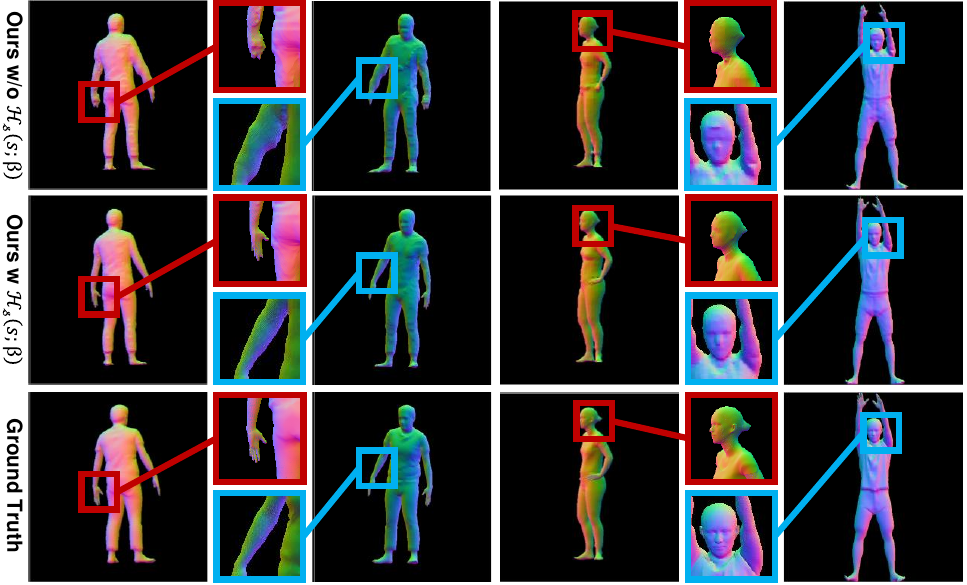}
    \caption{Reconstruction results with or without our progressive high-frequency SDF $\oursdf$. The geometry details of clothes, hands, faces are enhanced by introducing $\oursdf$.}
    \label{fig:ablation_hpsdf}
\end{figure}

\subsection{Toy Experiments}
\label{SEC:TOYEXP}
Our motivation stems from the idea that HF information and LF information improve details and robustness, respectively. To verify this, we employ two tools, \ie~ \textit{SDF} and \textit{voxelized SMPL-X} $\mathcal{M}_v^{3D}$ to establish this constraint. 

\textbf{Impact of SDF.} We build upon three representative methods, \ie~PIFu, PaMIR, and ICON for the experiments. %
Specifically, $\mathrm{PIFu}_{{w}~\mathrm{SDF}}$~adds SDF following equations: 
$\mathcal{\phi}(\mathcal{F}_{s}(\mathbf{p}), f_\text{\twoD}(\mathcal{I})({\mathbf{p}})) \rightarrow \hat{\mathcal{O}}$.
$\mathrm{PaMIR}_{{w}~\mathrm{SDF}}$ incoorporates SDF following:
$\mathcal{\phi}(\mathcal{F}_{s}(\mathbf{p}), f_\text{\twoD}(\mathcal{I})({\mathbf{p}}), \mathcal{V}(\bodyMesh)({\mathbf{p}}))\rightarrow \hat{\mathcal{O}}$.
$\mathrm{ICON}_{{w/o}~\mathrm{SDF}}$ replaces SDF with the z coordinate of $\mathbf{p}$ following:
    $\mathcal{\phi}(\mathcal{F}_{\mathrm{n}}^{\mathrm{b}}(\mathbf{p}), \mathcal{F}_{\mathrm{n}}^{\mathrm{c}}(\mathbf{p}), \mathbf{p}_z) \rightarrow \hat{\mathcal{O}}$, where $\mathcal{\phi}$ is the vanilla implicit function and $f_{2D}$ is a 2D CNN.
 Experimental results in Fig.~\ref{fig:motivation}, and Tab.~\ref{tab:TOY} demonstrate the SDF improves geometry details compared with variant methods without it.

\textbf{Impact of voxelized SMPL-X $M_v^{3D}$.} Our empirical verification is based on ICON, D-IF and \sexyname~that requires SMPL-X for reconstruction. we design three variants named $\mathrm{ICON}_{\mathrm{w}~\mathcal{M}_v^{3D}(p)}$, $\mathrm{D}$-$\mathrm{IF}_{\mathrm{w}~\mathcal{M}_v^{3D}(p)}$ and $\mathrm{\sexyname}_{\mathrm{w/o}~\mathcal{M}_v^{3D}(p)}$ that add or remove $M_v^{3D}$. 
From the experimental results in Fig.~\ref{fig:motivation} and Tab.~\ref{tab:smplnoise}, we find that incorporating $\mathcal{M}_v^{3D}$ helps achieve a more robust reconstruction even faces various levels of noise in SMPL-X shape and pose. The reason is that the low-resolution voxel grid of the naked body is insensitive to noise, and therefore provides robust low-frequency regularization in the training process.

\begin{figure}
    \centering
\includegraphics[width=0.5\textwidth]{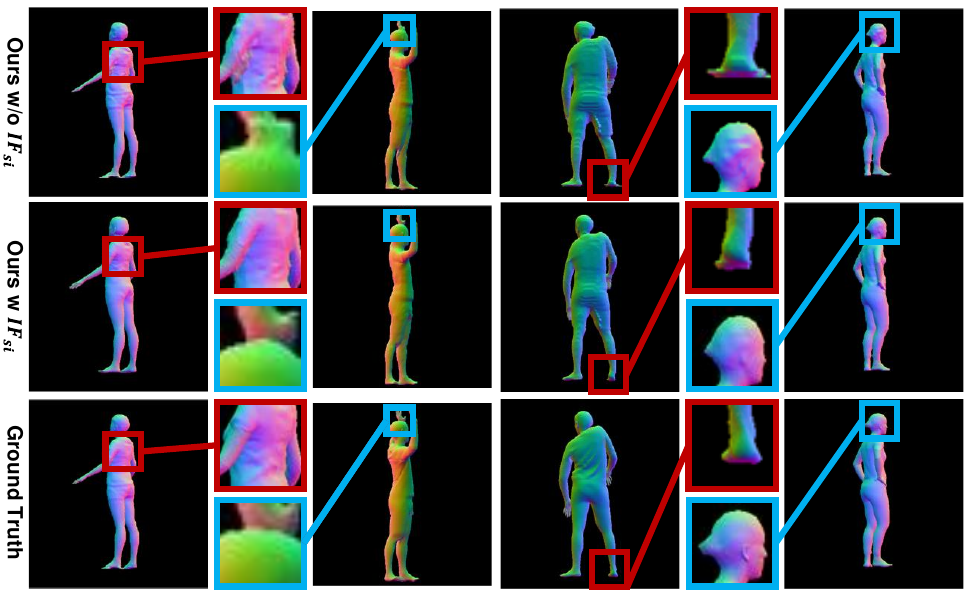}
    \caption{Reconstructions w and w/o our spatial interaction implicit function $\ourmlp$. Our $\ourmlp$ is able to perceive the global human body and is therefore able to remove non-human shapes.}
    \label{fig:ifsi}
\end{figure}
\subsection{Comparison Experiments}
\textbf{Quantitative results.} We conducted comparative experiments in Tab.~\ref{tab:comparison} under two settings. 1) \emph{Setting1}: Following the setting of the previous methods, we train and test on the same datasets.  2) \emph{Setting2}: To further evaluate the generalization ability of our \sexyname~on unseen datasets, we train and test \sexyname~using different datasets. Our approach achieves the best results in the seen and the unseen settings due to the high- and low-frequency paradigm.

\begin{figure}
    \centering
    \includegraphics[width=0.5\textwidth]{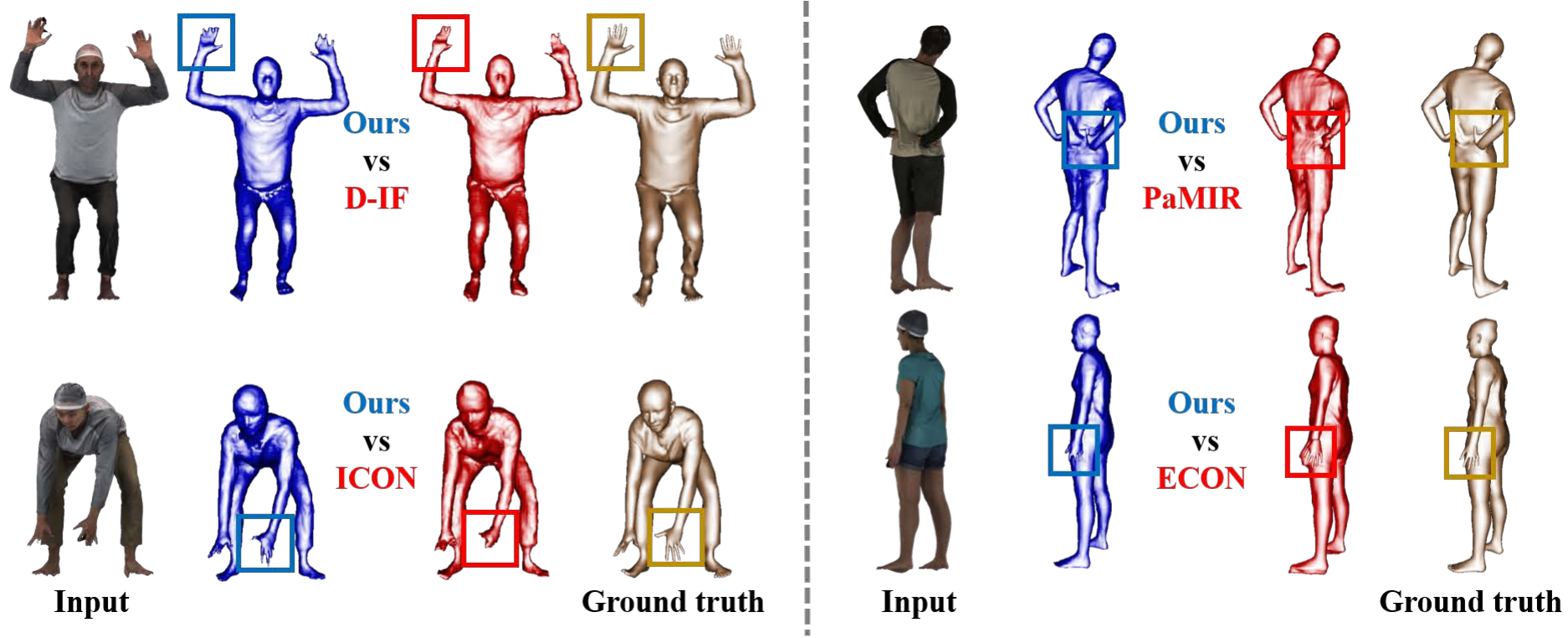}
    \caption{Visualization comparisons on CAPE dataset. The model is training on Thuman2.0 dataset.}
    \label{fig:vs_cape}
\end{figure}
\noindent \textbf{Visualization Results.}
We compare our \sexyname~with baselines on in-the-wild images and CAPE dataset in Fig.~\ref{fig:vis_comp}, and Fig.~\ref{fig:vs_cape}, respectively. 
The results show that our \sexyname~is able to reconstruct 3D clothed avatars with more realistic details. Although ECON obtains detailed fingers by replacing the hand of the SMPL-X model, there exists misalignment on the connection wrist when the corresponding SMPL-X is inaccurate.
We put more visualization results of our \sexyname~on in-the-wild images in Fig.~\ref{fig:inthewild}.
The results demonstrate the effectiveness and generalization ability of our \sexyname~in recovering detailed geometry (such as hairs and cloth wrinkles). 
We put more visualization results in the Appx.

\subsection{Ablation Studies} \label{sec:ablation}
\textbf{How does} $\mathbf{\oursdf}$ \textbf{improve geometry details?}
We quantitatively demonstrate the necessity of $\oursdf$~in Tab.~\ref{tab:comparison}. The results demonstrate that $\mathrm{\sexyname}_{{w/o}~\oursdf}$~, the variant method that replaces $\oursdf$ with standard SDF, achieves inferior performance than \sexyname. 
To further study the impact of $\oursdf$ on common (-FP) and challenging poses (-NFP), we evaluate \sexyname~on cape dataset that contains both categories.  Tab.~\ref{tab:comparison} demonstrates that $\oursdf$ improves the performance of avatar reconstruction more on challenging poses ($9.54\%$ improvement) than in fashion ($5.21\%$ improvement in terms of Chamfer distance. Furthermore, Fig.~\ref{fig:ablation_hpsdf} demonstrates that $\oursdf$~leads to more detailed reconstruction, resulting in clearer arms and more realistic cloth wrinkles. From the results, we observe that incorporating the power of high frequency with SDF helps in capturing detailed geometry. \\
\textbf{How does} $\mathbf{\ourmlp}$ \textbf{improve body topology of the reconstructed avatar?}
As shown in Fig.~\ref{fig:ifsi} and Tab.~\ref{tab:comparison}, our $\ourmlp$ removes the non-human shape and boosts reconstruction performance. The reason is that our $\ourmlp$ leverages a cross-scale attention module $\mathcal{A}$ that builds topological signals between different spatial points in the body model.

\subsection{Further Discussions}
\label{sec:discuss}
\noindent \textbf{Is our} $\mathbf{\mathcal{H}_{s}(p;\beta)}$ \textbf{able to help \sexyname~be robust to SMPL-X noise?} 
We study the impact of $\mathcal{H}_{s}(p;\beta)$ on the robustness ability of our \sexyname~by replacing it with conventional high frequency SDF $\mathcal{H}_{s}(p)$ and vanilla SDF. 
We perturb the SMPL-X model with various levels of noise to compare the robustness of the SDF variants. As illustrated in Tab.~\ref{tab:smplnoise_sdf}, our proposed $\mathcal{H}_{s}(p;\beta)$ outperforms SDF and high-frequency SDF variants under multiple noise scales due to the progressive manner. See more results in the Appx.

\noindent \textbf{Is \sexyname~able to converge faster?}
\begin{figure}
    \centering
    \includegraphics[width=0.45\textwidth]{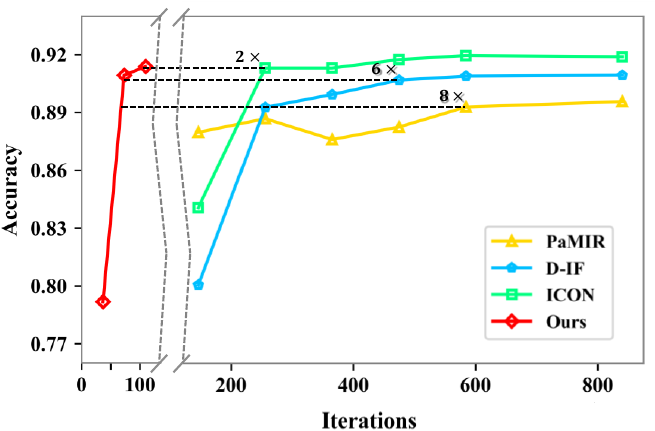}
    \caption{Convergence curves of different methods on CAPE dataset. Our \sexyname ~is able to converge faster than existing methods.}
    \label{fig:converge}
\end{figure}
In the comparison of validation accuracy depicted in Fig.~\ref{fig:converge}, it is evident that our \sexyname~ exhibits a remarkable ability to rapidly converge and attain superior performance. Specifically, \sexyname~swiftly reaches a commendable accuracy of 0.90 at approximately iteration 100. In contrast, the second best method, \ie~ICON, takes significantly longer, around iteration 200, to reach the same level of accuracy, underscoring the efficiency and efficacy of our approach. %

\section{Conclusion}
In this paper, we propose a high-frequency and low-frequency paradigm by exploiting high-frequency and low-frequency information from parametric body models.  Based on the paradigm, we design \sexynamefullname, namely 
\sexyname~that contains: 1) a progressive high-frequency SDF to improve geometry details and alleviate large gradients that hinder model convergence; 2) a spatial interaction implicit function that utilizes the low-frequency complementary information from the voxelized naked body to improve robustness against noise. Experimental results demonstrate the superiority of our \sexyname.
In the future, we will apply our method to more 3D reconstruction tasks such as 3D face reconstruction, and indoor scene 3D reconstruction.
\section{Acknowledgement}
 This work was partially supported by National Natural Science Foundation of China (NSFC) 62072190, Program for Guangdong Introducing Innovative and Entrepreneurial Teams 2017ZT07X183, and TCL Science and Technology Innovation Fund.

\clearpage
{
    \small
    \bibliographystyle{ieeenat_fullname}
    \bibliography{main}

\begin{thebibliography}{66}
\providecommand{\natexlab}[1]{#1}
\providecommand{\url}[1]{\texttt{#1}}
\expandafter\ifx\csname urlstyle\endcsname\relax
  \providecommand{\doi}[1]{doi: #1}\else
  \providecommand{\doi}{doi: \begingroup \urlstyle{rm}\Url}\fi

\bibitem[Alldieck et~al.(2019{\natexlab{a}})Alldieck, Magnor, Bhatnagar, Theobalt, and Pons-Moll]{alldieck2019learning}
Thiemo Alldieck, Marcus Magnor, Bharat~Lal Bhatnagar, Christian Theobalt, and Gerard Pons-Moll.
\newblock Learning to reconstruct people in clothing from a single rgb camera.
\newblock In \emph{Proceedings of the IEEE/CVF Conference on Computer Vision and Pattern Recognition}, pages 1175--1186, 2019{\natexlab{a}}.

\bibitem[Alldieck et~al.(2019{\natexlab{b}})Alldieck, Pons-Moll, Theobalt, and Magnor]{alldieck2019tex2shape}
Thiemo Alldieck, Gerard Pons-Moll, Christian Theobalt, and Marcus Magnor.
\newblock Tex2shape: Detailed full human body geometry from a single image.
\newblock In \emph{Proceedings of the IEEE/CVF International Conference on Computer Vision}, pages 2293--2303, 2019{\natexlab{b}}.

\bibitem[Allen-Zhu et~al.(2019)Allen-Zhu, Li, and Song]{allen2019convergence}
Zeyuan Allen-Zhu, Yuanzhi Li, and Zhao Song.
\newblock On the convergence rate of training recurrent neural networks.
\newblock \emph{Advances in neural information processing systems}, 32, 2019.

\bibitem[Anguelov et~al.(2005)Anguelov, Srinivasan, Koller, Thrun, Rodgers, and Davis]{anguelov2005scape}
Dragomir Anguelov, Praveen Srinivasan, Daphne Koller, Sebastian Thrun, Jim Rodgers, and James Davis.
\newblock Scape: shape completion and animation of people.
\newblock In \emph{ACM SIGGRAPH 2005 Papers}, pages 408--416. 2005.

\bibitem[Aseeri and Interrante(2021)]{aseeri2021influence}
Sahar Aseeri and Victoria Interrante.
\newblock The influence of avatar representation on interpersonal communication in virtual social environments.
\newblock \emph{IEEE transactions on visualization and computer graphics}, 27\penalty0 (5):\penalty0 2608--2617, 2021.

\bibitem[Bell et~al.(2016)Bell, Zitnick, Bala, and Girshick]{bell2016inside}
Sean Bell, C~Lawrence Zitnick, Kavita Bala, and Ross Girshick.
\newblock Inside-outside net: Detecting objects in context with skip pooling and recurrent neural networks.
\newblock In \emph{Proceedings of the IEEE conference on computer vision and pattern recognition}, pages 2874--2883, 2016.

\bibitem[Bhatnagar et~al.(2019)Bhatnagar, Tiwari, Theobalt, and Pons-Moll]{bhatnagar2019multi}
Bharat~Lal Bhatnagar, Garvita Tiwari, Christian Theobalt, and Gerard Pons-Moll.
\newblock Multi-garment net: Learning to dress 3d people from images.
\newblock In \emph{Proceedings of the IEEE/CVF international conference on computer vision}, pages 5420--5430, 2019.

\bibitem[Bhatnagar et~al.(2020{\natexlab{a}})Bhatnagar, Sminchisescu, Theobalt, and Pons-Moll]{bhatnagar2020combining}
Bharat~Lal Bhatnagar, Cristian Sminchisescu, Christian Theobalt, and Gerard Pons-Moll.
\newblock Combining implicit function learning and parametric models for 3d human reconstruction.
\newblock In \emph{The European Conference on Computer Vision}, pages 311--329, 2020{\natexlab{a}}.

\bibitem[Bhatnagar et~al.(2020{\natexlab{b}})Bhatnagar, Sminchisescu, Theobalt, and Pons-Moll]{bhatnagar2020loopreg}
Bharat~Lal Bhatnagar, Cristian Sminchisescu, Christian Theobalt, and Gerard Pons-Moll.
\newblock Loopreg: Self-supervised learning of implicit surface correspondences, pose and shape for 3d human mesh registration.
\newblock \emph{Advances in Neural Information Processing Systems}, 33:\penalty0 12909--12922, 2020{\natexlab{b}}.

\bibitem[Bu et~al.(2023)Bu, Huang, and Cui]{bu2023towards}
Qingwen Bu, Dong Huang, and Heming Cui.
\newblock Towards building more robust models with frequency bias.
\newblock In \emph{Proceedings of the IEEE/CVF International Conference on Computer Vision}, pages 4402--4411, 2023.

\bibitem[Chen et~al.(2022)Chen, Ren, and Yan]{chen2022rethinking}
Yiting Chen, Qibing Ren, and Junchi Yan.
\newblock Rethinking and improving robustness of convolutional neural networks: a shapley value-based approach in frequency domain.
\newblock In \emph{Advances in Neural Information Processing Systems}, pages 324--337, 2022.

\bibitem[Chen and Zhang(2019)]{chen2019learning}
Zhiqin Chen and Hao Zhang.
\newblock Learning implicit fields for generative shape modeling.
\newblock In \emph{Proceedings of the IEEE/CVF Conference on Computer Vision and Pattern Recognition}, pages 5939--5948, 2019.

\bibitem[Corona et~al.(2021)Corona, Pumarola, Alenya, Pons-Moll, and Moreno-Noguer]{corona2021smplicit}
Enric Corona, Albert Pumarola, Guillem Alenya, Gerard Pons-Moll, and Francesc Moreno-Noguer.
\newblock Smplicit: Topology-aware generative model for clothed people.
\newblock In \emph{Proceedings of the IEEE/CVF conference on computer vision and pattern recognition}, pages 11875--11885, 2021.

\bibitem[Dong et~al.(2022)Dong, Guo, Song, Chen, Geiger, and Hilliges]{dong2022pina}
Zijian Dong, Chen Guo, Jie Song, Xu Chen, Andreas Geiger, and Otmar Hilliges.
\newblock Pina: Learning a personalized implicit neural avatar from a single rgb-d video sequence.
\newblock In \emph{Proceedings of the IEEE/CVF Conference on Computer Vision and Pattern Recognition}, pages 20470--20480, 2022.

\bibitem[Eksombatchai et~al.(2018)Eksombatchai, Jindal, Liu, Liu, Sharma, Sugnet, Ulrich, and Leskovec]{eksombatchai2018pixie}
Chantat Eksombatchai, Pranav Jindal, Jerry~Zitao Liu, Yuchen Liu, Rahul Sharma, Charles Sugnet, Mark Ulrich, and Jure Leskovec.
\newblock Pixie: A system for recommending 3+ billion items to 200+ million users in real-time.
\newblock In \emph{Proceedings of the 2018 world wide web conference}, pages 1775--1784, 2018.

\bibitem[Galvane et~al.(2015)Galvane, Ronfard, Lino, and Christie]{galvane2015continuity}
Quentin Galvane, R{\'e}mi Ronfard, Christophe Lino, and Marc Christie.
\newblock Continuity editing for 3d animation.
\newblock In \emph{Proceedings of the AAAI Conference on Artificial Intelligence}, 2015.

\bibitem[Guo et~al.(2019)Guo, Lincoln, Davidson, Busch, Yu, Whalen, Harvey, Orts-Escolano, Pandey, Dourgarian, et~al.]{guo2019relightables}
Kaiwen Guo, Peter Lincoln, Philip Davidson, Jay Busch, Xueming Yu, Matt Whalen, Geoff Harvey, Sergio Orts-Escolano, Rohit Pandey, Jason Dourgarian, et~al.
\newblock The relightables: Volumetric performance capture of humans with realistic relighting.
\newblock \emph{ACM Transactions on Graphics}, 38\penalty0 (6):\penalty0 1--19, 2019.

\bibitem[He et~al.(2020)He, Collomosse, Jin, and Soatto]{he2020geo}
Tong He, John Collomosse, Hailin Jin, and Stefano Soatto.
\newblock Geo-pifu: Geometry and pixel aligned implicit functions for single-view human reconstruction.
\newblock \emph{Advances in Neural Information Processing Systems}, 33:\penalty0 9276--9287, 2020.

\bibitem[He et~al.(2021)He, Xu, Saito, Soatto, and Tung]{he2021arch++}
Tong He, Yuanlu Xu, Shunsuke Saito, Stefano Soatto, and Tony Tung.
\newblock Arch++: Animation-ready clothed human reconstruction revisited.
\newblock In \emph{Proceedings of the IEEE/CVF international conference on computer vision}, pages 11046--11056, 2021.

\bibitem[Huang et~al.(2019)Huang, Wang, Huang, Huang, Wei, and Liu]{huang2019ccnet}
Zilong Huang, Xinggang Wang, Lichao Huang, Chang Huang, Yunchao Wei, and Wenyu Liu.
\newblock Ccnet: Criss-cross attention for semantic segmentation.
\newblock In \emph{Proceedings of the IEEE/CVF international conference on computer vision}, pages 603--612, 2019.

\bibitem[Huang et~al.(2020)Huang, Xu, Lassner, Li, and Tung]{huang2020arch}
Zeng Huang, Yuanlu Xu, Christoph Lassner, Hao Li, and Tony Tung.
\newblock Arch: Animatable reconstruction of clothed humans.
\newblock In \emph{Proceedings of the IEEE/CVF Conference on Computer Vision and Pattern Recognition}, pages 3093--3102, 2020.

\bibitem[Jaderberg et~al.(2015)Jaderberg, Simonyan, Zisserman, et~al.]{jaderberg2015spatial}
Max Jaderberg, Karen Simonyan, Andrew Zisserman, et~al.
\newblock Spatial transformer networks.
\newblock \emph{Advances in neural information processing systems}, 28, 2015.

\bibitem[Jiang et~al.(2020)Jiang, Zhang, Hong, Luo, Liu, and Bao]{jiang2020bcnet}
Boyi Jiang, Juyong Zhang, Yang Hong, Jinhao Luo, Ligang Liu, and Hujun Bao.
\newblock Bcnet: Learning body and cloth shape from a single image.
\newblock In \emph{The European Conference on Computer Vision}, pages 18--35. Springer, 2020.

\bibitem[Jo et~al.(2016)Jo, Kim, and Kim]{jo2016effects}
Dongsik Jo, Ki-Hong Kim, and Gerard~Jounghyun Kim.
\newblock Effects of avatar and background representation forms to co-presence in mixed reality (mr) tele-conference systems.
\newblock In \emph{SIGGRAPH ASIA 2016 virtual reality meets physical reality: modelling and simulating virtual humans and environments}, pages 1--4. 2016.

\bibitem[Li et~al.(2024)Li, Chen, Wu, Yu, and Tan]{LI2024cross}
Daiyuan Li, Guo Chen, Xixian Wu, Zitong Yu, and Mingkui Tan.
\newblock Cross-stage relation extraction and presentation attack material perception for face anti-spoofing.
\newblock \emph{Neural Networks}, page 106275, 2024.

\bibitem[Li et~al.(2020)Li, Shen, Guo, and Lai]{li2020wavelet}
Qiufu Li, Linlin Shen, Sheng Guo, and Zhihui Lai.
\newblock Wavelet integrated cnns for noise-robust image classification.
\newblock In \emph{Proceedings of the IEEE/CVF Conference on Computer Vision and Pattern Recognition}, pages 7245--7254, 2020.

\bibitem[Li et~al.(2022)Li, Guillard, Remelli, and Fua]{li2022dig}
Ren Li, Beno{\^\i}t Guillard, Edoardo Remelli, and Pascal Fua.
\newblock Dig: Draping implicit garment over the human body.
\newblock In \emph{Proceedings of the Asian Conference on Computer Vision}, pages 2780--2795, 2022.

\bibitem[Loper et~al.(2015)Loper, Mahmood, Romero, Pons-Moll, and Black]{loper2015smpl}
Matthew Loper, Naureen Mahmood, Javier Romero, Gerard Pons-Moll, and Michael~J Black.
\newblock Smpl: A skinned multi-person linear model.
\newblock \emph{Acm Transactions on Graphics}, 34, 2015.

\bibitem[Lorensen and Cline(1998)]{lorensen1998marching}
William~E Lorensen and Harvey~E Cline.
\newblock Marching cubes: A high resolution 3d surface construction algorithm.
\newblock In \emph{Seminal graphics: pioneering efforts that shaped the field}, pages 347--353, 1998.

\bibitem[Ma et~al.(2020)Ma, Yang, Ranjan, Pujades, Pons-Moll, Tang, and Black]{CAPEma2020learning}
Qianli Ma, Jinlong Yang, Anurag Ranjan, Sergi Pujades, Gerard Pons-Moll, Siyu Tang, and Michael~J Black.
\newblock Learning to dress 3d people in generative clothing.
\newblock In \emph{Proceedings of the IEEE/CVF Conference on Computer Vision and Pattern Recognition}, pages 6469--6478, 2020.

\bibitem[Martin-Brualla et~al.(2021)Martin-Brualla, Radwan, Sajjadi, Barron, Dosovitskiy, and Duckworth]{martin2021nerf}
Ricardo Martin-Brualla, Noha Radwan, Mehdi~SM Sajjadi, Jonathan~T Barron, Alexey Dosovitskiy, and Daniel Duckworth.
\newblock Nerf in the wild: Neural radiance fields for unconstrained photo collections.
\newblock In \emph{Proceedings of the IEEE/CVF Conference on Computer Vision and Pattern Recognition}, pages 7210--7219, 2021.

\bibitem[Mescheder et~al.(2019)Mescheder, Oechsle, Niemeyer, Nowozin, and Geiger]{mescheder2019occupancy}
Lars Mescheder, Michael Oechsle, Michael Niemeyer, Sebastian Nowozin, and Andreas Geiger.
\newblock Occupancy networks: Learning 3d reconstruction in function space.
\newblock In \emph{Proceedings of the IEEE/CVF conference on computer vision and pattern recognition}, pages 4460--4470, 2019.

\bibitem[Mildenhall et~al.(2020)Mildenhall, Srinivasan, Tancik, Barron, Ramamoorthi, and Ng]{mildenhall2020nerf}
Ben Mildenhall, Pratul~P. Srinivasan, Matthew Tancik, Jonathan~T. Barron, Ravi Ramamoorthi, and Ren Ng.
\newblock Nerf: Representing scenes as neural radiance fields for view synthesis.
\newblock In \emph{The European Conference on Computer Vision}, 2020.

\bibitem[Moon et~al.(2022)Moon, Nam, Shiratori, and Lee]{moon20223d}
Gyeongsik Moon, Hyeongjin Nam, Takaaki Shiratori, and Kyoung~Mu Lee.
\newblock 3d clothed human reconstruction in the wild.
\newblock In \emph{European conference on computer vision}, pages 184--200, 2022.

\bibitem[Newell et~al.(2016)Newell, Yang, and Deng]{newell2016stacked}
Alejandro Newell, Kaiyu Yang, and Jia Deng.
\newblock Stacked hourglass networks for human pose estimation.
\newblock In \emph{The European Conference on Computer Vision}, pages 483--499, 2016.

\bibitem[Noh et~al.(2015)Noh, Yeo, and Woo]{noh2015hmd}
Seung-Tak Noh, Hui-Shyong Yeo, and Woontack Woo.
\newblock An hmd-based mixed reality system for avatar-mediated remote collaboration with bare-hand interaction.
\newblock In \emph{Proceedings of the 25th International Conference on Artificial Reality and Telexistence and 20th Eurographics Symposium on Virtual Environments}, pages 61--68, 2015.

\bibitem[Park et~al.(2019)Park, Florence, Straub, Newcombe, and Lovegrove]{park2019deepsdf}
Jeong~Joon Park, Peter Florence, Julian Straub, Richard Newcombe, and Steven Lovegrove.
\newblock Deepsdf: Learning continuous signed distance functions for shape representation.
\newblock In \emph{Proceedings of the IEEE/CVF conference on computer vision and pattern recognition}, pages 165--174, 2019.

\bibitem[Park et~al.(2021)Park, Sinha, Barron, Bouaziz, Goldman, Seitz, and Martin-Brualla]{park2021nerfies}
Keunhong Park, Utkarsh Sinha, Jonathan~T. Barron, Sofien Bouaziz, Dan~B Goldman, Steven~M. Seitz, and Ricardo Martin-Brualla.
\newblock Nerfies: Deformable neural radiance fields.
\newblock \emph{Proceedings of the IEEE/CVF International Conference on Computer Vision}, pages 5865--5874, 2021.

\bibitem[Pascanu et~al.(2013)Pascanu, Mikolov, and Bengio]{pascanu2013difficulty}
Razvan Pascanu, Tomas Mikolov, and Yoshua Bengio.
\newblock On the difficulty of training recurrent neural networks.
\newblock In \emph{International conference on machine learning}, pages 1310--1318, 2013.

\bibitem[Paszke et~al.(2019)Paszke, Gross, Massa, Lerer, Bradbury, Chanan, Killeen, Lin, Gimelshein, Antiga, et~al.]{paszke2019pytorch}
Adam Paszke, Sam Gross, Francisco Massa, Adam Lerer, James Bradbury, Gregory Chanan, Trevor Killeen, Zeming Lin, Natalia Gimelshein, Luca Antiga, et~al.
\newblock Pytorch: An imperative style, high-performance deep learning library.
\newblock \emph{Advances in neural information processing systems}, 32, 2019.

\bibitem[Pavlakos et~al.(2019)Pavlakos, Choutas, Ghorbani, Bolkart, Osman, Tzionas, and Black]{pavlakos2019smplx}
Georgios Pavlakos, Vasileios Choutas, Nima Ghorbani, Timo Bolkart, Ahmed~AA Osman, Dimitrios Tzionas, and Michael~J Black.
\newblock Expressive body capture: 3d hands, face, and body from a single image.
\newblock In \emph{Proceedings of the IEEE/CVF conference on computer vision and pattern recognition}, pages 10975--10985, 2019.

\bibitem[Rahaman et~al.(2019{\natexlab{a}})Rahaman, Baratin, Arpit, Draxler, Lin, Hamprecht, Bengio, and Courville]{rahaman2019bias}
Nasim Rahaman, Aristide Baratin, Devansh Arpit, Felix Draxler, Min Lin, Fred Hamprecht, Yoshua Bengio, and Aaron Courville.
\newblock On the spectral bias of neural networks.
\newblock In \emph{International Conference on Machine Learning}, pages 5301--5310. PMLR, 2019{\natexlab{a}}.

\bibitem[Rahaman et~al.(2019{\natexlab{b}})Rahaman, Baratin, Arpit, Draxler, Lin, Hamprecht, Bengio, and Courville]{rahaman2019spectral}
Nasim Rahaman, Aristide Baratin, Devansh Arpit, Felix Draxler, Min Lin, Fred Hamprecht, Yoshua Bengio, and Aaron Courville.
\newblock On the spectral bias of neural networks.
\newblock In \emph{International Conference on Machine Learning}, pages 5301--5310, 2019{\natexlab{b}}.

\bibitem[Saito et~al.(2019)Saito, Huang, Natsume, Morishima, Kanazawa, and Li]{saito2019pifu}
Shunsuke Saito, Zeng Huang, Ryota Natsume, Shigeo Morishima, Angjoo Kanazawa, and Hao Li.
\newblock Pifu: Pixel-aligned implicit function for high-resolution clothed human digitization.
\newblock In \emph{Proceedings of the IEEE/CVF international conference on computer vision}, pages 2304--2314, 2019.

\bibitem[Saito et~al.(2020)Saito, Simon, Saragih, and Joo]{saito2020pifuhd}
Shunsuke Saito, Tomas Simon, Jason Saragih, and Hanbyul Joo.
\newblock Pifuhd: Multi-level pixel-aligned implicit function for high-resolution 3d human digitization.
\newblock In \emph{Proceedings of the IEEE/CVF Conference on Computer Vision and Pattern Recognition}, pages 84--93, 2020.

\bibitem[Sch{\"o}nberger et~al.(2016)Sch{\"o}nberger, Zheng, Frahm, and Pollefeys]{schonberger2016pixelwise}
Johannes~L Sch{\"o}nberger, Enliang Zheng, Jan-Michael Frahm, and Marc Pollefeys.
\newblock Pixelwise view selection for unstructured multi-view stereo.
\newblock In \emph{The European Conference on Computer Vision}, 2016.

\bibitem[Sitzmann et~al.(2020)Sitzmann, Martel, Bergman, Lindell, and Wetzstein]{sitzmann2020siren}
Vincent Sitzmann, Julien Martel, Alexander Bergman, David Lindell, and Gordon Wetzstein.
\newblock Implicit neural representations with periodic activation functions.
\newblock \emph{Advances in neural information processing systems}, 33:\penalty0 7462--7473, 2020.

\bibitem[Sun et~al.(2022)Sun, Sun, and Chen]{sun2022voxelgrid}
Cheng Sun, Min Sun, and Hwann-Tzong Chen.
\newblock Direct voxel grid optimization: Super-fast convergence for radiance fields reconstruction.
\newblock In \emph{Proceedings of the IEEE/CVF Conference on Computer Vision and Pattern Recognition}, pages 5459--5469, 2022.

\bibitem[Sun(2022)]{sun2022research}
Lin Sun.
\newblock Research on the application of 3d animation special effects in animated films: Taking the film avatar as an example.
\newblock \emph{Scientific Programming}, 2022.

\bibitem[Tieleman et~al.(2012)Tieleman, Hinton, et~al.]{tieleman2012lecture}
Tijmen Tieleman, Geoffrey Hinton, et~al.
\newblock Lecture 6.5-rmsprop: Divide the gradient by a running average of its recent magnitude.
\newblock \emph{COURSERA: Neural networks for machine learning}, 4\penalty0 (2):\penalty0 26--31, 2012.

\bibitem[Willumsen(2018)]{willumsen2018my}
Ea~Christina Willumsen.
\newblock Is my avatar my avatar? character autonomy and automated avatar actions in digital games.
\newblock In \emph{DiGRA Conference}, 2018.

\bibitem[Xiang et~al.(2020)Xiang, Prada, Wu, and Hodgins]{xiang2020monoclothcap}
Donglai Xiang, Fabian Prada, Chenglei Wu, and Jessica Hodgins.
\newblock Monoclothcap: Towards temporally coherent clothing capture from monocular rgb video.
\newblock In \emph{2020 International Conference on 3D Vision (3DV)}, pages 322--332. IEEE, 2020.

\bibitem[Xiu et~al.(2022)Xiu, Yang, Tzionas, and Black]{xiu2022icon}
Yuliang Xiu, Jinlong Yang, Dimitrios Tzionas, and Michael~J. Black.
\newblock {ICON}: {I}mplicit {C}lothed humans {O}btained from {N}ormals.
\newblock In \emph{Proceedings of the IEEE/CVF Conference on Computer Vision and Pattern Recognition}, pages 13296--13306, 2022.

\bibitem[Xiu et~al.(2023)Xiu, Yang, Cao, Tzionas, and Black]{xiu2023econ}
Yuliang Xiu, Jinlong Yang, Xu Cao, Dimitrios Tzionas, and Michael~J. Black.
\newblock {ECON: Explicit Clothed humans Optimized via Normal integration}.
\newblock In \emph{Proceedings of the IEEE/CVF Conference on Computer Vision and Pattern Recognition}, 2023.

\bibitem[Yang et~al.(2023{\natexlab{a}})Yang, Luo, Xiu, Wang, Xu, and Fan]{yang2023dif}
Xueting Yang, Yihao Luo, Yuliang Xiu, Wei Wang, Hao Xu, and Zhaoxin Fan.
\newblock D-if: Uncertainty-aware human digitization via implicit distribution field.
\newblock In \emph{Proceedings of the IEEE/CVF International Conference on Computer Vision}, pages 9122--9132, 2023{\natexlab{a}}.

\bibitem[Yang et~al.(2023{\natexlab{b}})Yang, Zhang, Huang, Zhang, and Tan]{yang2023cross}
Yifan Yang, Shuhai Zhang, Zixiong Huang, Yubing Zhang, and Mingkui Tan.
\newblock Cross-ray neural radiance fields for novel-view synthesis from unconstrained image collections.
\newblock In \emph{Proceedings of the IEEE/CVF International Conference on Computer Vision}, pages 15901--15911, 2023{\natexlab{b}}.

\bibitem[Yoon et~al.(2019)Yoon, Kim, Lee, Billinghurst, and Woo]{yoon2019effect}
Boram Yoon, Hyung-il Kim, Gun~A Lee, Mark Billinghurst, and Woontack Woo.
\newblock The effect of avatar appearance on social presence in an augmented reality remote collaboration.
\newblock In \emph{2019 IEEE Conference on Virtual Reality and 3D User Interfaces (VR)}, pages 547--556. IEEE, 2019.

\bibitem[Zhang et~al.(2021)Zhang, Tian, Zhou, Ouyang, Liu, Wang, and Sun]{zhang2021pymaf}
Hongwen Zhang, Yating Tian, Xinchi Zhou, Wanli Ouyang, Yebin Liu, Limin Wang, and Zhenan Sun.
\newblock Pymaf: 3d human pose and shape regression with pyramidal mesh alignment feedback loop.
\newblock In \emph{Proceedings of the IEEE/CVF International Conference on Computer Vision}, pages 11446--11456, 2021.

\bibitem[Zhang et~al.(2023{\natexlab{a}})Zhang, Tian, Zhang, Li, An, Sun, and Liu]{zhang2023pymafx}
Hongwen Zhang, Yating Tian, Yuxiang Zhang, Mengcheng Li, Liang An, Zhenan Sun, and Yebin Liu.
\newblock Pymaf-x: Towards well-aligned full-body model regression from monocular images.
\newblock \emph{IEEE Transactions on Pattern Analysis and Machine Intelligence}, 2023{\natexlab{a}}.

\bibitem[Zhang et~al.(2023{\natexlab{b}})Zhang, Rao, and Agrawala]{zhang2023adding}
Lvmin Zhang, Anyi Rao, and Maneesh Agrawala.
\newblock Adding conditional control to text-to-image diffusion models.
\newblock In \emph{Proceedings of the IEEE Conference on Computer Vision and Pattern Recognition}, 2023{\natexlab{b}}.

\bibitem[Zhang and Zhu(2019)]{zhang2019interpreting}
Tianyuan Zhang and Zhanxing Zhu.
\newblock Interpreting adversarially trained convolutional neural networks.
\newblock In \emph{International conference on machine learning}, pages 7502--7511, 2019.

\bibitem[Zheng et~al.(2014)Zheng, Dunn, Jojic, and Frahm]{zheng2014patchmatch}
Enliang Zheng, Enrique Dunn, Vladimir Jojic, and Jan-Michael Frahm.
\newblock Patchmatch based joint view selection and depthmap estimation.
\newblock In \emph{Proceedings of the IEEE Conference on Computer Vision and Pattern Recognition}, pages 1510--1517, 2014.

\bibitem[Zheng et~al.(2019)Zheng, Yu, Wei, Dai, and Liu]{zheng2019deephuman}
Zerong Zheng, Tao Yu, Yixuan Wei, Qionghai Dai, and Yebin Liu.
\newblock Deephuman: 3d human reconstruction from a single image.
\newblock In \emph{Proceedings of the IEEE/CVF International Conference on Computer Vision}, pages 7739--7749, 2019.

\bibitem[Zheng et~al.(2021)Zheng, Yu, Liu, and Dai]{zheng2021pamir}
Zerong Zheng, Tao Yu, Yebin Liu, and Qionghai Dai.
\newblock Pamir: Parametric model-conditioned implicit representation for image-based human reconstruction.
\newblock \emph{IEEE transactions on pattern analysis and machine intelligence}, 44\penalty0 (6):\penalty0 3170--3184, 2021.

\bibitem[Zhu et~al.(2019)Zhu, Zuo, Wang, Cao, and Yang]{zhu2019detailed}
Hao Zhu, Xinxin Zuo, Sen Wang, Xun Cao, and Ruigang Yang.
\newblock Detailed human shape estimation from a single image by hierarchical mesh deformation.
\newblock In \emph{Proceedings of the IEEE/CVF conference on computer vision and pattern recognition}, pages 4491--4500, 2019.

\bibitem[Zixiong et~al.(2024)Zixiong, Qi, Libo, Yifan, Naizhou, Mingkui, and Qi]{huang2024g}
Huang Zixiong, Chen Qi, Sun Libo, Yang Yifan, Wang Naizhou, Tan Mingkui, and Wu Qi.
\newblock G-nerf: Geometry-enhanced novel view synthesis from single-view images.
\newblock \emph{arXiv preprint arXiv:2404.07474}, 2024.

\end{thebibliography}
}

\clearpage
\setcounter{page}{1}
\maketitlesupplementary

\setcounter{section}{0}
\setcounter{equation}{0}
\setcounter{table}{0}
\setcounter{figure}{0}
\renewcommand\thesection{\Alph{section}}
\renewcommand{\thetable}{\Alph{table}}
\renewcommand{\thefigure}{\Alph{figure}}
\renewcommand{\theequation}{\Alph{equation}}

\etocdepthtag.toc{mtappendix}
\etocsettagdepth{mtchapter}{none}
    \etocsettagdepth{mtappendix}{subsection}
    
    {
        \hypersetup{linkcolor=black}
    	\footnotesize\tableofcontents
    }
\section{Related Works}

\textbf{Explicit-shape-based approaches} rely on parametric human body models, \eg~SCAPE~\cite{anguelov2005scape}, SMPL~\cite{loper2015smpl}, SMPL-X~\cite{pavlakos2019smplx} to reconstruct 3D humans. Many works~\cite{alldieck2019learning, alldieck2019tex2shape, CAPEma2020learning, xiang2020monoclothcap, zhu2019detailed, bhatnagar2019multi} introduce the concept of "body+offset", where clothing geometry is represented as 3D displacements on top of the SMPL models. For example, MGN~\cite{bhatnagar2019multi} proposes a top-down objective function to align the segmentation maps of predicted garments and SMPL. %
To improve the expression ability of garment templates and support more topologies, BCNet~\cite{jiang2020bcnet} disentangles the skinning weight of the garment from the body mesh. Different from the representation of "body+offset", alternative parametric methods adapt vertex deformations on body mesh to capture cloth details. For example, HMD~\cite{zhu2019detailed} presents the hierarchical deformation framework to recover a detailed human body shape from an initial SMPL mesh in a coarse-to-fine manner. The advantage of these methods lies in their compatibility with the current animation pipeline and ease of control through pose parameters. However, they have limitations in modeling various and complex clothing topologies due to the inherent topology constraints imposed by parametric models. 

\textbf{Implicit-function-based approaches} aims to reconstruct detailed surfaces with arbitrary topology~\cite{chen2019learning, mescheder2019occupancy, park2019deepsdf}. This is achieved through the implicit functions, which can be used to approximate 3D representation such as occupancy fields or signed distance fields. PIFu~\cite{saito2019pifu} is the pioneering method that utilizes pixel-aligned features for the regression of the occupancy field of human shape. PIFuHD~\cite{saito2020pifuhd} incorporates a multi-level architecture and additional normals to improve the geometric details of PIFu. However, these two methods lack constraints on the global topology of humans, leading to performance degradation in challenging poses. Many works attempt to address this issue in different ways, such as introducing a coarse shape of volumetric humans~\cite{he2020geo}, leveraging depth information of RGB-D images~\cite{dong2022pina}. Unlike the above methods, alternative implicit-function-based methods learn the latent representation of clothing to control the generation of clothing~\cite{moon20223d, li2022dig, corona2021smplicit}. For example, SMPLicit~\cite{corona2021smplicit} reconstructs the clothed human by optimizing the latent space of the clothing model to control clothing cut and style. However, the reconstructed human still does not align well with the input image and lacks geometric details.

\textbf{Explicit shape \& Implicit function approaches} leverage human body models and implicit functions to harness the benefits of both worlds~\cite{bhatnagar2020combining, bhatnagar2020loopreg, huang2020arch}. For instance, PaMIR~\cite{zheng2021pamir} regularizes the free-form implicit function by incorporating semantic features from the SMPL model. ICON\cite{xiu2022icon}, on the other hand, regresses shapes from locally queried features to generalize to unseen poses in in-the-wild photos. ECON~\cite{xiu2023econ} combines estimated 2.5D front and back surfaces with an underlying 3D parametric body for improved reconstruction. To further address the variations in distribution among different spatial points, D-IF~\cite{yang2023dif} introduces a distribution to express the uncertainty of clothing. 
However, these approaches may fall short when performing highly detailed and robust reconstruction. Specifically, PaMIR is sensitive to global pose and lacks robustness to unseen poses~\cite{xiu2022icon}. ECON is prone to reconstruct combined or broken limbs due to the need to complete different surfaces. ICON and D-IF tend to fail in reconstructing detailed parts such as elbows and wrinkles in clothes (see Fig. 1.%
To promote a detailed reconstruction, our \sexyname~ uses a progressive high-frequency function to improve the expression of reconstructed details. At the same time, \sexyname~ uses the low-frequency-based spatial interactive implicit function to enhance robustness to unseen shapes and poses.

\section{More Details of our \sexyname}
\subsection{Novelty and differences from previous methods ~\cite{he2020geo, saito2020pifuhd, xiu2022icon, zheng2021pamir}} Clothed human reconstruction from a single RGB image is challenging due to limited views and the absence of depth information. Most recent methods ~\cite{xiu2022icon, xiu2023econ, zheng2021pamir, yang2023dif} rely on parametric body models estimated from RGB images,
but they may incur an oversmooth problem
due to the \textbf{underutilization} of high-frequency (HF) details.
Moreover, these methods can be \textbf{sensitive} to noises incurred by parametric body model estimation for challenging poses.\\
To address the above issues, we first \textbf{enhance} HF information from the body models to describe geometry details. To this end, we design a progressive growing function to achieve accurate reconstruction while alleviating the convergence difficulty associated with HF information. Moreover, we verify low-frequency (LF) information from the parametric model is \textbf{insensitive} to noise. Considering this, we establish a spatial interaction function to leverage the (LF) for robustness reconstruction. 
\subsection{Details on Spatial Interaction MLP}
\label{sec_appx:detail_simlp}
\begin{figure}[t]
    \centering
    \includegraphics[width=0.45\textwidth]{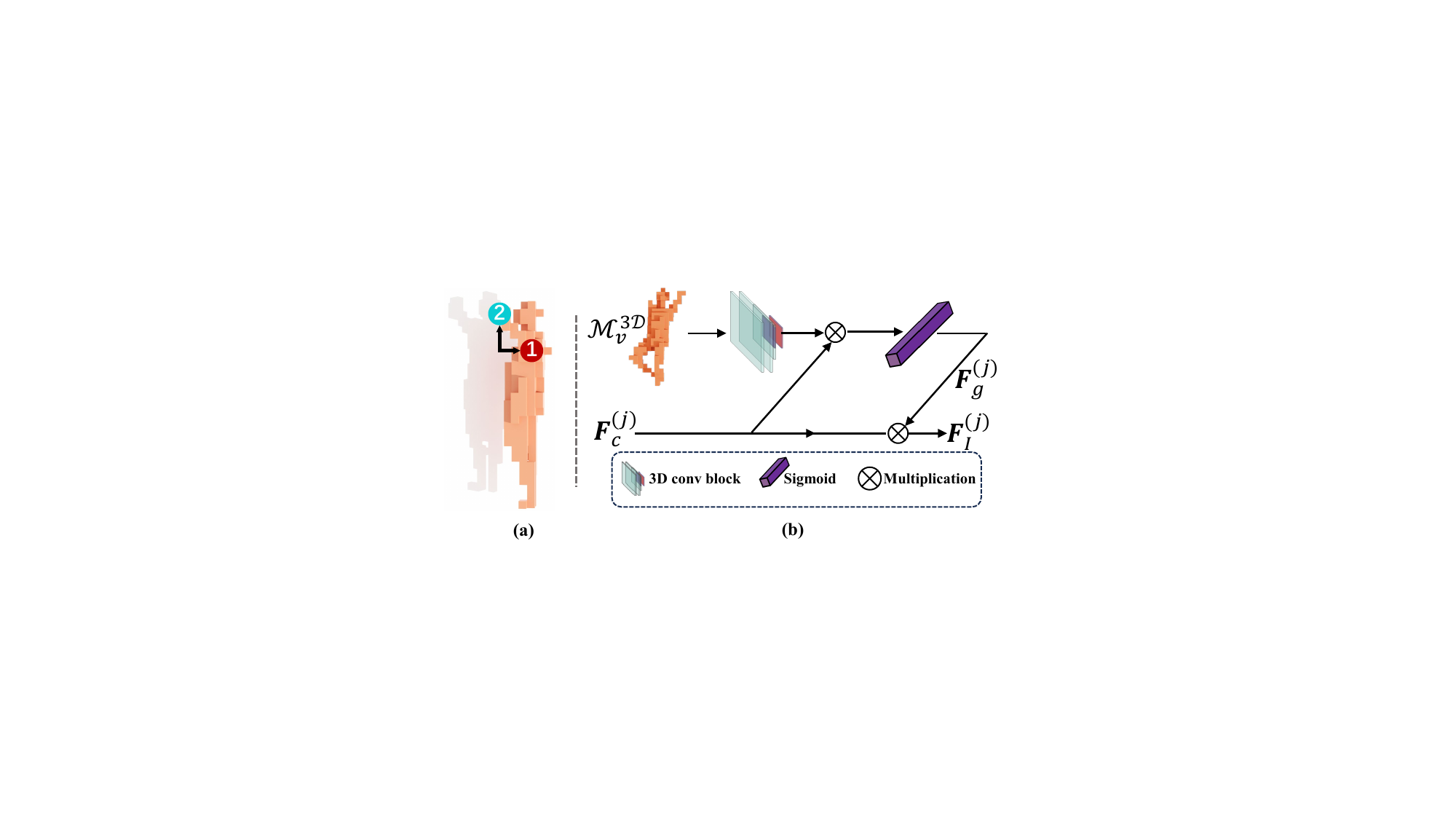}
    \vspace{-4pt}
    \caption{Illustration of the spatial interaction module $\mathcal{A}$.}
    \label{fig:simlp_only}
\end{figure}
Our spatial interaction implicit function takes $\mathbf{F}_{c}^1$ that contains our high-frequency SDF $\mathcal{H}(s;\beta)$, low-frequency voxel grids feature $\bodyMesh^{3D}_{\mathcal{V}}({\mathbf{p}})$, and normal features $\mathbf{F}_{\mathrm{n}}(\mathbf{p})$ as input and infers occupancy fields $\mathcal{\hat{O}}$. 
\begin{align} \label{eqn:ourmlp}
&\mathbf{F}_{c}^1=[\mathcal{H}(s;\beta), \bodyMesh^{3D}_{\mathcal{V}}({\mathbf{p}}), \mathbf{F}_{\mathrm{n}}(\mathbf{p})]\\
    &{\phi_{si}}(\mathbf{F}_{c}^1) \to \mathcal{\hat{O}},~~\phi_{si}(\cdot)=\mathcal{A}^{N+1} \circ T^{(N+1)} \circ  \cdots \circ \mathcal{A}^{1}(\cdot) \circ T^{(1)}.
\end{align}
As shown in Fig.~\ref{fig:simlp_only}, take the $1$-th layer of $\phi_{si}$ as an example,  we use attention module~\cite{huang2019ccnet, LI2024cross} $\mathcal{A}^{1}$ that takes in the $\bodyMesh^{3D}_{\mathcal{V}}$ and
$\mathbf{F}_{c}^{(1)}$ and output a spatial interaction feature map $\mathbf{F}_{I}^{(1)}$.
Specifically, We first extract a global spatial features $\mathbf{F}_{g}^{(j)}$ of the 
$\ourvoxel$ via a 3D Convolution block and a sigmoid function.
We achieve the spatial interaction process of different voxels through the equation $\mathbf{F}_{c}^{(j)}\times \mathbf{F}_{g}^{(j)} \rightarrow \mathbf{F}_{I}^{(j)}$. 
After obtaining the $\mathbf{F}_{I}^{(1)}$, we fed it to the first full-connected layer $T^{(1)}$ to obtain $\mathbf{F}^{2}_{c}$.

\subsection{Future work and limitations.}
\noindent\textbf{Q5. Future work and limitations.}
Since \sexyname~is trained on orthographic views, it struggles with strong perspectives, causing asymmetrical limbs or unrealistic shapes. This issue is worth studying in the future. \\

\section{More Experimental Details}  
 \begin{algorithm}[t]
 \small 
	\caption{The inference pipeline of \sexyname.}
	\label{alg:inference_hilo}
	\KwIn{Sampled 3D points $\{\mathbf{p}\}_{i=1}^{n}$, an RGB image $\mathcal{I}$ of human, an spatial interactive implicit function $\ourmlp$, a parametric body model estimation net $E_{p}$, a progressive high frequency function $\mathcal{H}(\cdot~;\beta)$, a 3D CNN $f_{3D}$, a mesh voxelization operation $\mathcal{V}$, a marching cubes operation $\mathcal{MC}$.}
 \KwOut{Triangular mesh of the human.}

  Obtaining parametric body model SMPL-X $\mathcal{M}$ with  $E_{p}(\mathcal{I})$.\\
 
		With $\mathcal{M}$, obtaining the global voxel grid $\ourvoxel$ using $f_{3D}(\mathcal{V}(\mathcal{M}))$.\\
 \For{$\mathbf{p}_{i}$ in $\{\mathbf{p}\}_{i=1}^{n}$}{
        Generating SDF $s$ \wrt~$\mathbf{p}_{i}$ using Eqn. 1.%

        Using $\mathcal{H}(\cdot~;\beta)$ to enhance the SDF $s$ resulting in point-wise progressive high-frequency SDF $\oursdf$. \\
        
        Obtaining the local voxel grid of $\mV$ by indexing $\ourvoxel$ with $\mathbf{p}_{i}$, resulting in $\mV(\mathbf{p}_{i})$.\\
        
        Get 3D normal features $\mathbf{F}_{\mathrm{n}}(\mathbf{p}_{i})$  \wrt~$\mathbf{p}_{i}$ following ICON.\\

        Concatenate $\oursdf$, $\mV(\mathbf{p})$, $\mathbf{F}_{\mathrm{n}}(\mathbf{p}_{i})$, getting $F_{c}^{1}$.\\

        Using $\ourmlp$ to obtaining occupancy field $\hat{\mO}(\mathbf{p}_{i})$ from $F_{c}^{1}$ and $\ourvoxel$, following Eqn. (7).\\

    }
            Obtaining the triangular mesh of the human using marching cubes algorithm with $\mathcal{MC}(\hat{\mO})$.
\end{algorithm}

We demonstrate the inference details of our \sexyname~in Alg.~\ref{alg:inference_hilo}. The 3D point set is obtained via a coarse-to-fine manner as illustrated in Sec.~\ref{sec:3dpoints}.

\subsection{Implementation Details}
Especially, the dimension of HFSDF, batch size $\mathrm{b}$, sampled points number $\mathrm{n}$, and variable dimension channels $\mathrm{C}$ of the spatial interaction module are set to $10$, $2$, $8000$, $[39, 512, 256, 128, 1]$ respectively. 
The training and testing phases are performed on a single NVIDIA GeForce RTX 3090 GPU.
See more details on the training and inference of \sexyname~in the appendix.

\subsection{More details on Metrics}
\label{sec_supp:metric}
Specifically, \textbf{P2S} denotes the distance between randomly sampled points from a ground truth mesh to its nearest surface on a reconstructed mesh. \textbf{Chamfer} is regarded as a bidirectional P2S distance, which computes the distance between randomly sampled points from the reconstructed mesh to its nearest surface on the ground truth mesh. \textbf{Normals} is calculated by measuring L2 error between normal images rendered from reconstructed and ground-truth meshes from fixed viewpoints. 
 
\subsection{3D Points Sampling} \label{sec:3dpoints}
During training, we randomly query 3D points inside, outside, and around the SMPL-X surface. During inference, we define the coordinates of 3D points through an initial 3D grid, and iteratively interpolate the 3D grid to sample 3D points in a more detailed scale.
\subsection{Details of Variant Methods}
\label{sec_appx:baseline}
\subsubsection{Revisit Existing Methods}  %
\noindent \textbf{PIFu}. To reconstruct a 3D-clothed human, PIFu proposes Pixel-Aligned Implicit Functions to predict whether each 3D point is inside or outside a human surface. Specifically, PIFu learns a 2D feature map from a single image $I$ using a 2D image encoder via $f_\text{\twoD}(\mathcal{I}) \rightarrow \mathcal{F}_{I}^{2D}$. To query local pixel-aligned features on $\mathcal{F}_{I}^{2D}$, PIFu projects 3D points $\mathbf{p}$ to a 2D plane with $\pi$ operation and uses bilinear interpolation operation $\mathrm{S}$ to sample the local features from $\mathcal{F}_{I}^{2D}$. 
The local feature $f_\text{\twoD}(\mathcal{I})(\mathbf{p})$ and the Z coordinate of $\mathbf{p}$ (\ie~ $\mathbf{p}_z$) are concatenated and fed to a multi-layer perceptron (MLP) to obtain the final prediction $\hat{\mathcal{O}}$. The pipeline of PIFU follows an equation:
\begin{equation}
    \mathrm{PIFu}   :\phi(f_\text{\twoD}(\mathcal{I})(\mathbf{p})), \mathbf{p}_z) \rightarrow \hat{\mathcal{O}}(\mathbf{p}))
\end{equation}
where 
$f_\text{\twoD}$    denotes the \twoD  image encoder. 
Although PIFu is able to reconstruct high-quality human mesh for commonly seen poses such as walking and standing, PIFu often fails when encountering severe occlusions and large pose variations due to insufficient information from a single image only. \\
\textbf{PaMIR.} To further regulate the reconstruction process, PaMIR introduces the strengths of parametric body models by learning a parametric-aligned 3D feature volume acquired from a parametric body model, \ie~ SMPL. Specifically, PaMIR estimates a SMPL model $\bodyMesh$ from the given single image $I$, converting $\bodyMesh$ to occupancy volume with mesh voxelization $\mathcal{V}$  and encoding the volume with 3D convolutional neural networks $f_\text{\threeD}$. Given the  voxel-aligned volume features $f_\text{\threeD}(\mathcal{V}(\bodyMesh)),{\mathbf{p}}))$ and the corresponding pixel-aligned feature vector $f_\text{\twoD}(\mathcal{I})({\mathbf{p}}))$ of $\mathbf{p}$, PaMIR learns an implicit function to predict whether $\mathbf{p}$ is inside or outside a human surface. The pipeline of PaMIR follows the equation:
\begin{equation}
\mathrm{PaMIR} : \phi((f_\text{\twoD}(\mathcal{I})({\mathbf{p}})), \ourvoxel({\mathbf{p}}))\rightarrow \hat{\mathcal{O}}(\mathbf{p}))
\end{equation}
Although PaMIR typically feeds their implicit-function module with features of a global 2D image or 3D voxel encoder, but these features are sensitive to global pose \cite{xiu2022icon}. \\
\textbf{ICON.} To improve the robustness to out-of-distribution poses, ICON replaces the global encoder of existing methods with a more local scheme: using signed distance function (SDF), barycentric surface normal and local normal features of SMPL regarding $\mathbf{p}$. The pipeline of ICON follows the equation:
\begin{equation}
\mathrm{ICON} : \phi(s(\mathbf{p}), \mathcal{F}_{\mathrm{n}}) \rightarrow \hat{\mathcal{O}}(\mathbf{p}))
\end{equation}
where $\mathcal{F}_{s}(\mathbf{p})$ is the signed distance 
from a query point $\mathbf{p}$ to the closest body point $\text{P}^\body \in \bodyMesh$, 
\highlight{and} 
$\bodyNormFeat$ is the barycentric surface normal of 
$\text{P}^\body$,
and $\cloNormFeat$ is a normal vector. We denote the concatenation of $\mathcal{F}_{\mathrm{n}}^{\mathrm{b}}(\mathbf{p})$, $\mathcal{F}_{\mathrm{n}}^{\mathrm{c}}(\mathbf{p})$ as $\mathcal{F}_{\mathrm{n}}$. 
\\
\textbf{D-IF.} To alleviate the uncertainty in the process of reconstructing a clothed human, D-IF follows ICON to estimate the occupancy field of the clothed human based on the equation:
\begin{equation}
\begin{aligned}
       \mathrm{D\text{-}IF}:~ &\hat{\mO}_f = \hat{\mO}_c + \phi_r(\hat{\mO}_c \oplus \mF_{7 \mathrm{D}} \oplus P_\varphi(\mF_{7 \mathrm{D}}))\\
       &\mF_{7D}=s \oplus \mathcal{F}_{\mathrm{n}},~ \hat{\mO}_c=\phi(\mF_{7 \mathrm{D}}) \\
        & \hat{\mO}_c(p) \sim P_{\varphi}\left(F_{7 \mathrm{D}}(\bp)\right)=\mathcal{N}\left(\mu_{\varphi}(\bp), \sigma_{\varphi}(\bp)\right)
\end{aligned}
\end{equation}
where $\oplus$ denotes concatenate operation, $P_{\varphi}(F_{7 \mathrm{D}}(\bp)$ is a Gaussian distribution.

\subsection{Variant Methods}
Based on the grasp of existing methods, we introduce the variant methods in our experiments.
\begin{equation}
\begin{aligned}
     &\mathrm{ICON}_{\mathrm{w}~\mathcal{M}_v^{3D}(p)}:  \phi(s(\bp), \mathcal{F}_{\mathrm{n}}, \ourvoxel(\bp)) \rightarrow \hat{\mathcal{O}}(\mathbf{p}))\\
  &\mathrm{D\text{-}IF}_{\mathrm{w}~\mathcal{M}_v^{3D}(p)}: \phi(\mF_{7 \mathrm{D}}, \ourvoxel(\bp))+ \phi_r(\hat{\mO}_c \oplus \mF_{7 \mathrm{D}} \oplus P_\varphi(\mF_{7 \mathrm{D}})) \\
  &\mathrm{\sexyname}_{\mathrm{w/o}~\mathcal{M}_v^{3D}(p)}: \ourmlp(\oursdf, \mathcal{F}_{\mathrm{n}}) \rightarrow \hat{\mathcal{O}}(\mathbf{p})) \\
   & \mathrm{\sexyname}_{\mathrm{w/o}~\mathcal{H}_s(p;\beta)} :  \ourmlp(s(\bp), \mathcal{F}_{\mathrm{n}}, \ourvoxel(\bp)) \rightarrow \hat{\mathcal{O}}(\mathbf{p})) \\
    & \mathrm{\sexyname}_{\mathrm{w}~\mathcal{H}_s(p)} : \ourmlp(\mH(s), \mathcal{F}_{\mathrm{n}}, \ourvoxel(\bp)) \rightarrow \hat{\mathcal{O}}(\mathbf{p}))  \\
    &\mathrm{\sexyname}_{\mathrm{w/o}~ \mathcal{\phi}_{si}}: \phi(\oursdf, \mathcal{F}_{\mathrm{n}}, \ourvoxel(\bp)) \rightarrow \hat{\mathcal{O}}(\mathbf{p}))\\
   & \mathrm{\sexyname}_{\mathrm{w/o}~ \mathcal{H}(s;\beta)~\mathrm{w/o}~ \mathcal{\phi}_{si}}: \phi(s(\bp), \mathcal{F}_{\mathrm{n}}, \ourvoxel(\bp)) \rightarrow \hat{\mathcal{O}}(\mathbf{p}))
\end{aligned}
\end{equation}

\section{More Details on Datasets}
Our data-split configuration aligns with the protocols outlined by ICON and D-IF. We conduct experiments on the basis of two distinct settings.
\begin{itemize}
    \item Setting 1: Train on Thuman2.0, test on CAPE.
For this setting, we employ 500 scans from Thuman2.0 for training, accompanied by 5 scans for validation. To assess reconstruction accuracy on CAPE, we utilize 150 scans, further categorized into challenging poses ("CAPE-NFP" - 100 scans) and fashion poses ("CAPE-FP" - 50 scans). To emulate diverse viewpoints during testing, RGB images are synthesized by rotating a virtual camera around the textured scans at angles of {$0^{\circ}, 120^{\circ}, 240^{\circ}$}.
\item Setting 2: Train and test on the same dataset.
In this scenario, when training and testing on Thuman2.0, we employ 500 scans for training and reserve 20 scans for testing. Conversely, when training and testing on CAPE, we utilize 120 scans for training, 5 for validation, and 25 for testing.
\end{itemize}
\section{More Experiments}
\subsection{More results on SMPL-X noise.}
\noindent \textbf{SMPL-X Model}.
Skinned Multi-Person Linear-Expressive model (SMPL-X)~\cite{pavlakos2019smplx} represents human body shapes and poses in a compact and parametric manner. The core idea behind SMPL is to use a linear combination of body shape parameters and joint rotations to represent a 3D human body model with $N{=}10475$ vertices and $K{=}54$ joints.  Specifically,  SMPL-X is defined by $\mathcal{M}(\theta, \beta, \psi): \mathbb{R}^{|\theta| \times |\beta| \times |\psi|} \to \mathbb{R}^{3N}$, where $\theta \in \mathbb{R}^{3(K+1)}$ represents the pose parameter, $\beta \in \mathbb{R}^{|\theta|}$ is the shape parameter, and $\psi$ denotes the facial expression parameters, and $K$ denotes the number of body joints in addition to a joint for global rotation. By adjusting $\theta$, $\beta$, $\psi$, SMPL-X is able to represent a wide variety of human body shapes and poses. See~\cite{pavlakos2019smplx} for more details.\\

\noindent \textbf{Adding noise to SMPL-X Model}.
We further evaluate the robustness ability of our \sexyname~against various levels of noise in the shape parameters $\theta_s$ and pose parameters $\theta_p$ in parametric models. Our experimental setting follows ICON, which samples a scalar value $\mu ~ \sim \mathcal{N}(0,1)$, scaling the noise with two predefined parameters $s_1$, $s_2$ to represent various levels of noise.
The above procedure follows the equation:
\begin{equation}
\begin{aligned}
    \theta_s+=s_1 \ast \mu \\
    \theta_p+=s_2 \ast \mu
\end{aligned}
\end{equation}
We set $\{s_1, s_2\}$ to $\{0.1, 0.1\}$, $\{0.2, 0.2\}$, $\{0.3, 0.3\}$, $\{0.4, 0.4\}$, $\{0.5, 0.5\}$ for a thorough study on the robustness of our \sexyname~and our baselines. Since we have provided the results \wrt~ $\{s_1, s_2\} \in [\{0.1, 0.1\}, \{0.2, 0.2\}, \{0.5, 0.5\}]$ in the main draft, we report the remaining results in Tab.~\ref{tab:more_noise_level} 

\begin{table*}
    \centering
    \begin{tabular}{l|c|cccccc}
    \toprule
    \cmidrule{1-8}    \multicolumn{1}{r}{} & \multicolumn{1}{r}{} & \multicolumn{3}{c}{SMPL-X Noise=0.3} & \multicolumn{3}{c}{SMPL-X Noise=0.4} \\
    \cmidrule{3-8}
    \multicolumn{1}{l}{Methods} & \multicolumn{1}{l}{$\mathcal{M}_v^{3D}$} & \multicolumn{1}{l}{CAPE-FP } & \multicolumn{1}{l}{CAPE-NFP } & \multicolumn{1}{l}{CAPE } & \multicolumn{1}{l}{CAPE-FP } & \multicolumn{1}{l}{CAPE-NFP} & \multicolumn{1}{l}{ CAPE } \\

    \midrule
    ICON  & \xmark & 4.5134  & 4.7091  & 4.7069  & 5.5864  & 5.9810  & 5.9015  \\
    ICON w $\mathcal{M}_v^{3D}$ & \cmark     & 4.2250  & 4.1215  & 4.2697  & 3.3824  & 3.4722  & 3.3897  \\
    \midrule
    D-IF  & \xmark     & 3.2462  & 3.6933  & 3.5700  & 3.2462  & 3.6933  & 3.5700  \\
    D-IF w $\mathcal{M}_v^{3D}$ & \cmark     & 1.2912  & 1.8222  & 1.5995  & 1.2912  & 1.8222  & 1.5995  \\
    \midrule
    \sexyname~w/o $\mathcal{M}_v^{3D}$ & \xmark     & 3.7060  & 4.3281  & 4.1071  & 4.4435  & 4.8639  & 4.7763  \\
    \sexyname & \cmark     & 1.1014  & 1.5407  & 1.3552  & 1.1633  & 1.7584  & 1.5132  \\
    \bottomrule
    \end{tabular}%
    \caption{Impact of $\ourvoxel$ on different methods in terms of Chamfer Distance. We train the models on Thuman2.0 and test them on CAPE.}
    \label{tab:more_noise_level}
\end{table*}
 
\subsection{More error measurements to assess robustness.} To further evaluate the robustness of our \sexyname, we calculate \textit{Chamfer}, \textit{P2S} and \textit{Normals} between SMPL-X and reconstructed body models. From Tab.~\ref{tab:robustness}, our \sexyname~shows better robustness than existing methods.
\\
   \captionof{table}
   {\footnotesize{Robustness on CAPE.}}
\setlength{\tabcolsep}{3pt} 
    \begin{threeparttable}
    \centering
\resizebox{0.7\linewidth}{!}{
{\Huge
\begin{tabular}{cccc}
    \toprule
         Methods & Chamfer ($\downarrow$)  & P2S ($\downarrow$)   & Normals ($\downarrow$)  \\
    \midrule
    PIFu  & 4.0550  & 3.3971  & 0.1915  \\
        PIFuHD & 6.1345 & 5.2692 & 0.2017 \\
    PaMIR & 0.9800  & 1.0132  & 0.0714  \\
    ICON  & 0.8198  & 0.7799  & 0.0617  \\
    \text{D-IF}  & 0.9111  & 0.8751  & 0.0666  \\
    \text{ECON}  & 0.9083  & 0.8701  & 0.0723  \\
    \sexyname~(Ours) & \textbf{0.6784}  & \textbf{0.6580}  &\textbf{0.0480}  \\
    \bottomrule
    \end{tabular}
    }%
    }
    \end{threeparttable}
 \label{tab:robustness}%

\subsection{Is \sexyname~efficient and light-weighted?}
\textbf{Comparison of inference and training time.}
In Tab.~\ref{tab:comp_speed}, we compare the inference efficiency by the average inference time to reconstruct 200 single-view images. The inference procedures of PaMIR, ICON, D-IF, and \sexyname~consist of SMPL-X fitting and cloth refinement. Differently, PIFu's inference procedure only includes cloth refinement, and ECON includes SMPL-X fitting and Poisson Surface Reconstruction (PSR). In terms of inference efficiency, it is evident that our \sexyname~ demonstrates a competitive performance with PaMIR, ICON and D-IF. However, ECON depends on time-consuming PSR to complete human shape, and all other methods show superior performance to it when inference. We measure training efficiency by the average time spent on 10 epochs on the Thuman2 dataset. D-IF needs to train two MLPs and therefore takes more time. We achieve competitive training efficiency with PIFu, PaMIR and ICON even though we introduce high-frequency and low-frequency information simultaneously. ECON lacks this statistic because the authors do not release the training codes.
\textbf{Comparison of model size.} Form Tab.~\ref{tab:comp_speed}, with the exception of ECON, the model sizes of existing methods are basically the same. Although ECON is lightweight, it requires time-consuming PSR to complete meshes of human shape.  
\begin{table*}[tbp]
\setlength\tabcolsep{4pt}
 \begin{center}
 \caption{\label{tab:comp_speed}Comparing training/inference efficiency and model size of existing methods.}
 \begin{tabular}{c|c|c|c}
 \toprule
  Method & Inference Time (seconds) & Training Time (seconds) & Million Parameters (seconds)\\
 \midrule
 PIFu~\cite{saito2019pifu} & 8.13 & 1636 & 28.09 \\
 PaMIR~\cite{zheng2021pamir} & 21.97 & 1298 & 28.18 \\
 ICON~\cite{xiu2022icon}  & 18.63 & 1697 & 28.11 \\
 D-IF~\cite{yang2023dif} & 18.51 & 2336 & 28.79\\
 ECON~\cite{xiu2023econ} & 110.93 &  - & 12.07\\
 \sexyname~(Ours) & 19.17 & 1918 & 28.21\\

 \bottomrule
 \end{tabular}
 \end{center}
\end{table*}

\section{More visualization Results}
\subsection{Transfer Sketch to 3D model}
\begin{figure*}
    \centering
    \includegraphics[width=1\linewidth]{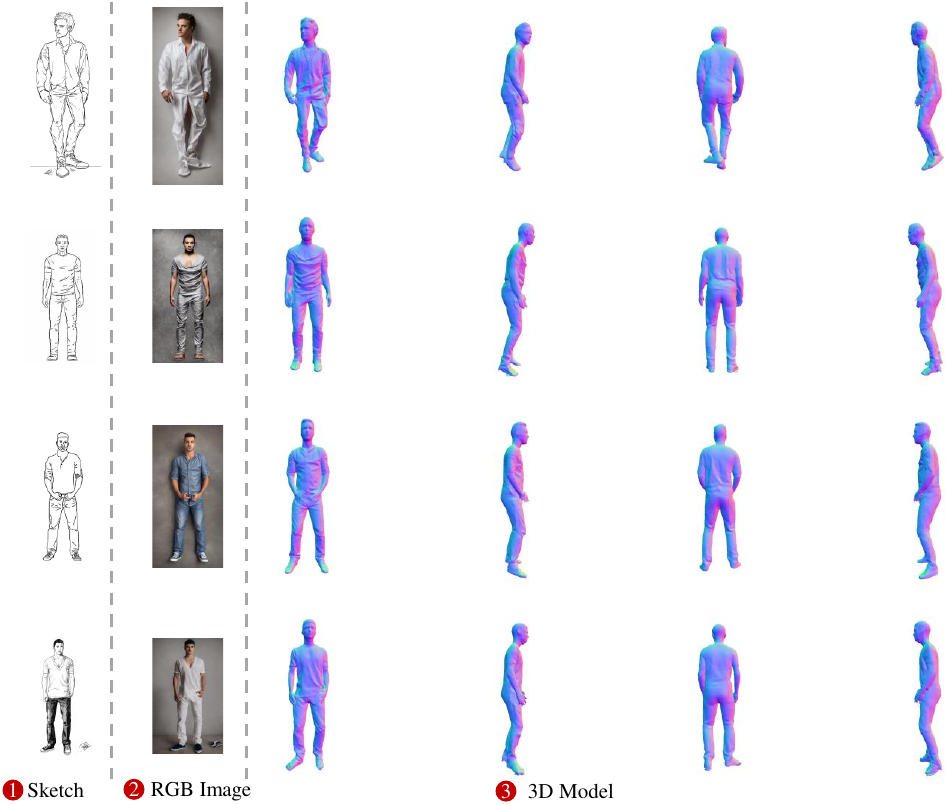}
    \caption{More application of our \sexyname. We are able to transfer a sketch of a clothed human into a 3D model.}
    \label{fig:sketch23d}
\end{figure*}

Since our \sexyname~is robust to in-the-wild images~\cite{yang2023cross, martin2021nerf}, we are able to put it to more applications. We show in Fig.~\ref{fig:sketch23d} that our \sexyname~is able to transfer a sketch image of a clothed human into a 3D model with the help of ControlNet~\cite{zhang2023adding}. Specifically, we collect sketch images from Pinterest and use ControlNet to transfer the images to RGB images. The RGB images are then fed to our \sexyname~to reconstruct 3D model of the corresponding human. \\
\subsection{Results on In-the-wild Images}
\begin{figure*}
    \centering
    \includegraphics[width=0.90\linewidth]{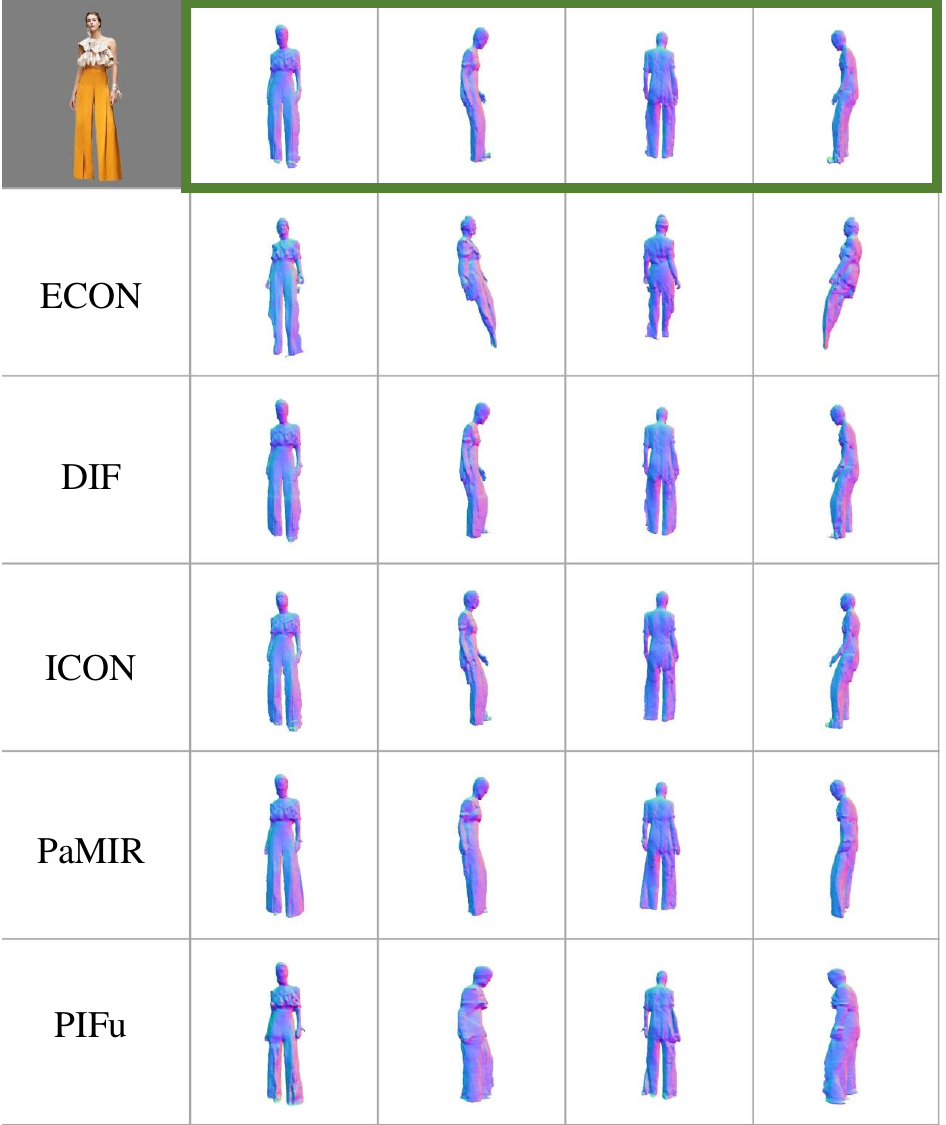}
    \caption{Visualization comparisons of reconstruction for our \textcolor[RGB]{84,130,53}{\sexyname~} vs SOTA.}
    \label{fig:sup_comp1_1}
\end{figure*}

\begin{figure*}
    \centering
    \includegraphics[width=0.90\linewidth]{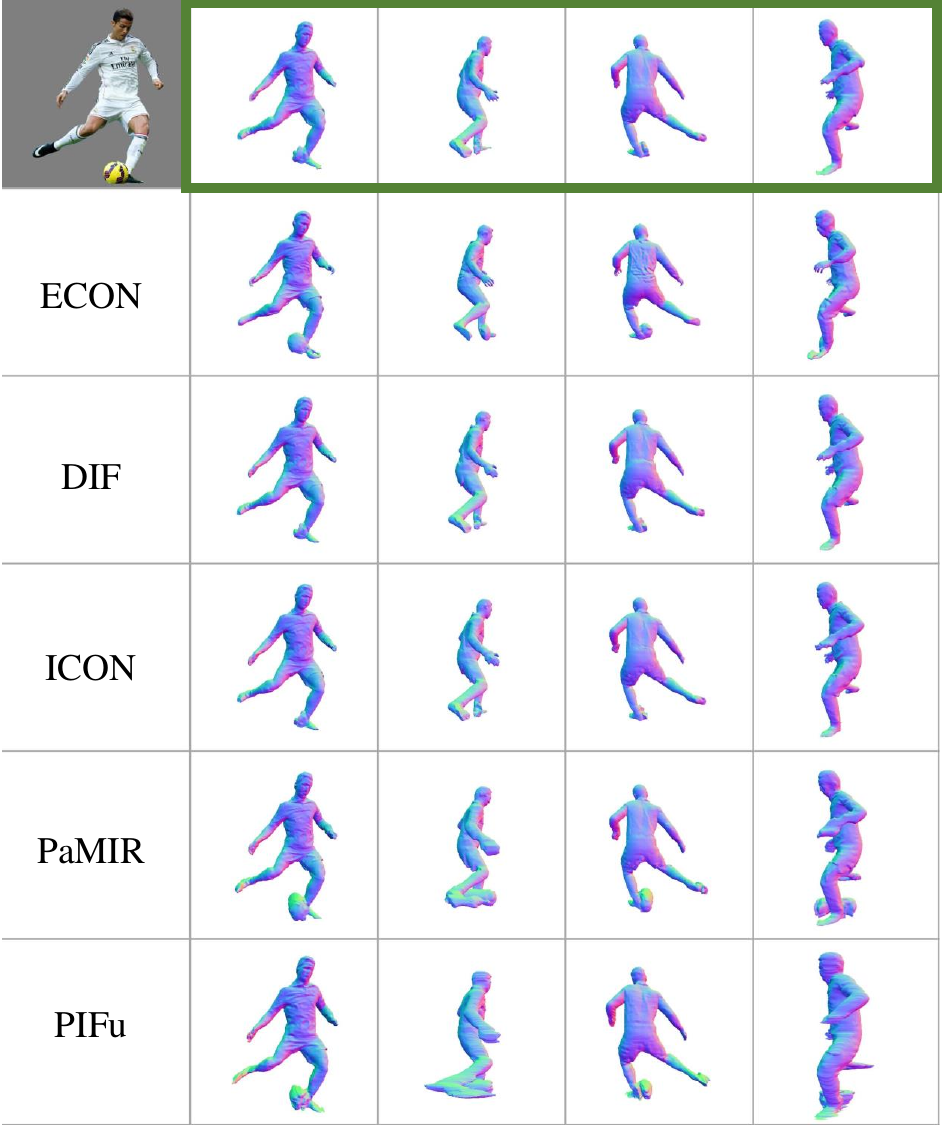}
    \caption{Visualization comparisons of reconstruction for our \textcolor[RGB]{84,130,53}{\sexyname~} vs SOTA.}
    \label{fig:sup_comp1_2}
\end{figure*}

\begin{figure*}
    \centering
    \includegraphics[width=0.90\linewidth]{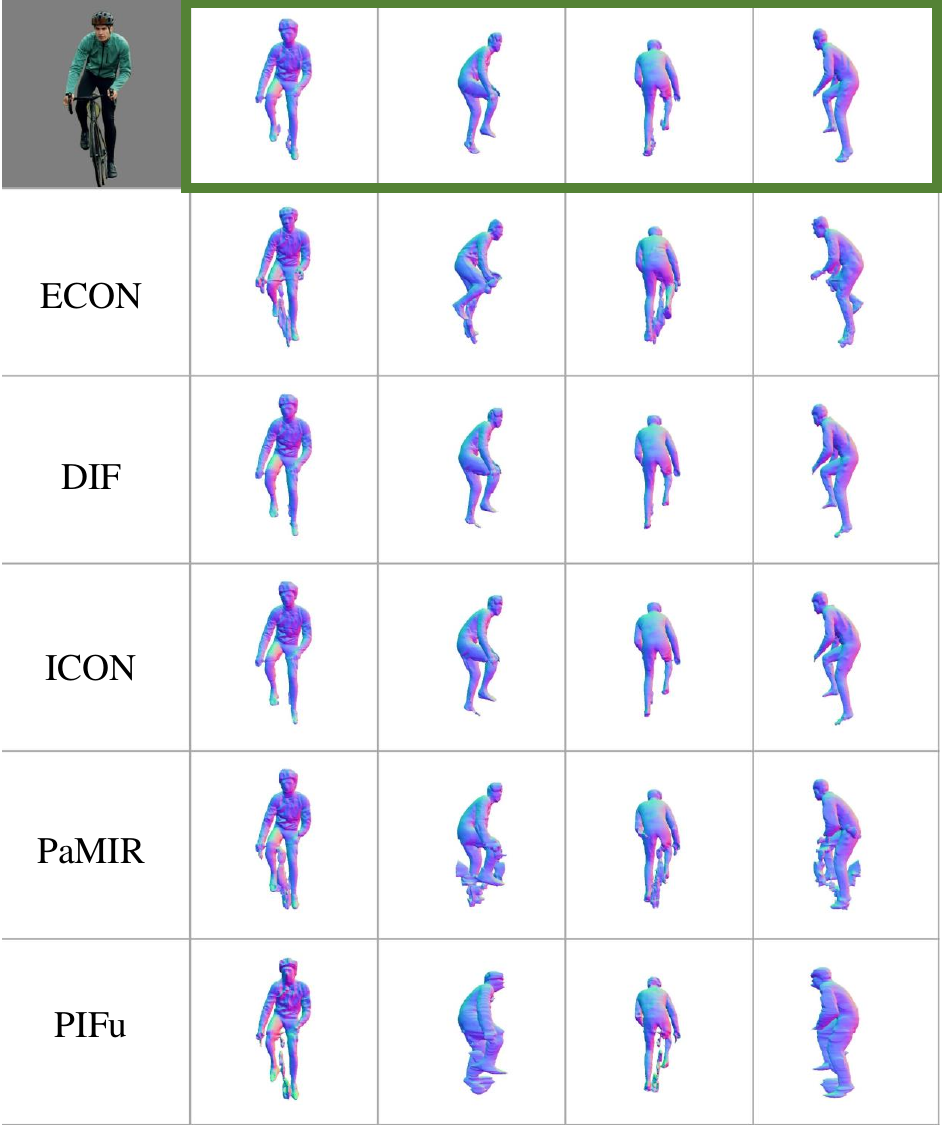}
    \caption{Visualization comparisons of reconstruction for our \textcolor[RGB]{84,130,53}{\sexyname~} vs SOTA.}
    \label{fig:sup_comp1_3}
\end{figure*}

\begin{figure*}
    \centering
    \includegraphics[width=0.90\linewidth]{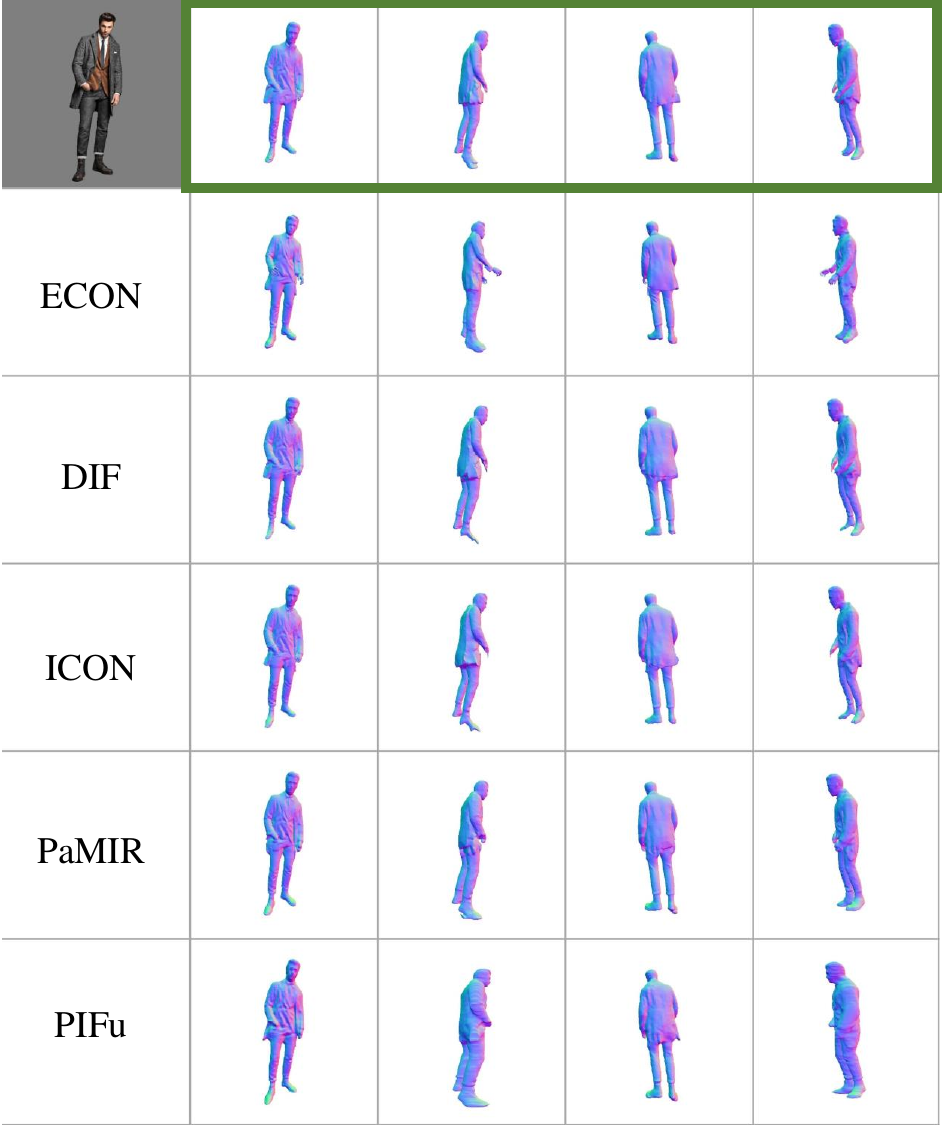}
    \caption{Visualization comparisons of reconstruction for our \textcolor[RGB]{84,130,53}{\sexyname~} vs SOTA.}
    \label{fig:sup_comp1_4}
\end{figure*}

\begin{figure*}
    \centering
    \includegraphics[width=0.90\linewidth]{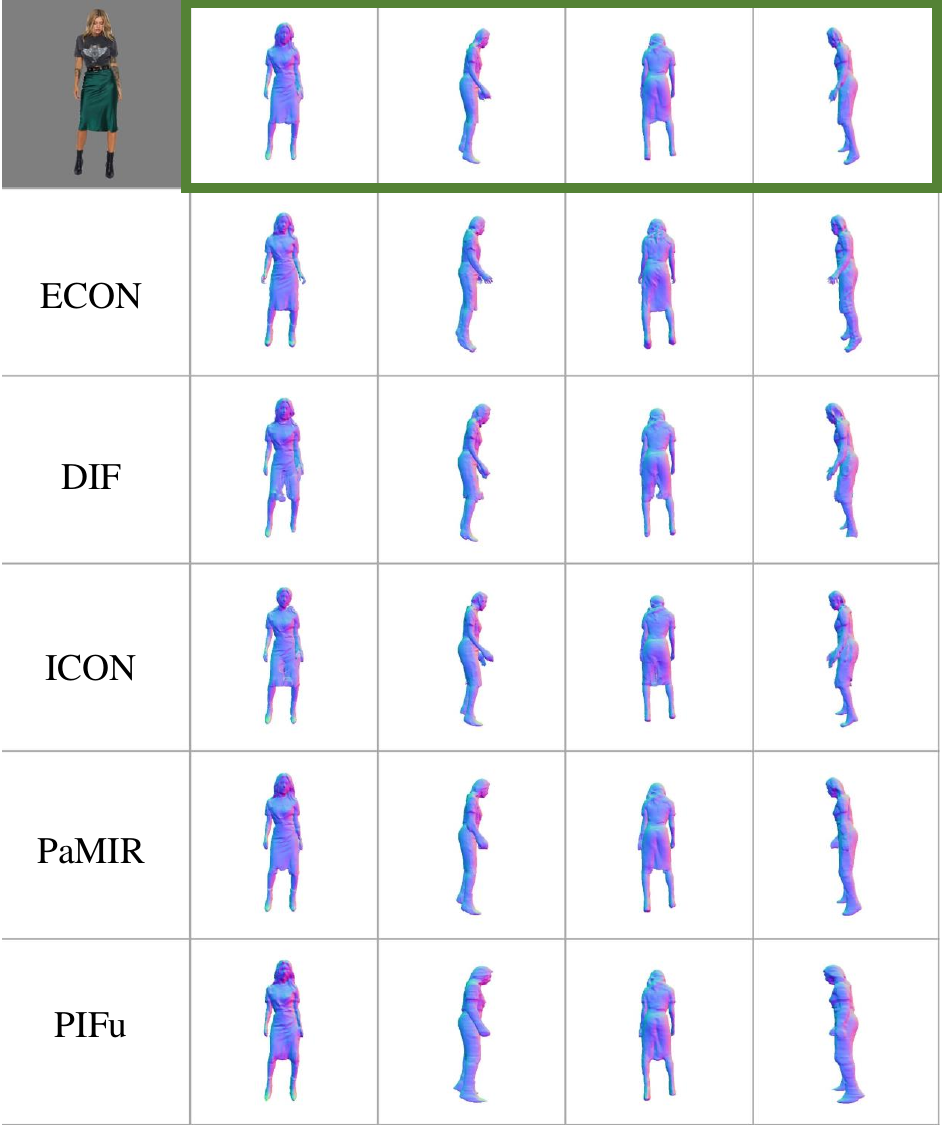}
    \caption{Visualization comparisons of reconstruction for our \textcolor[RGB]{84,130,53}{\sexyname~} vs SOTA.}
    \label{fig:sup_comp1_5}
\end{figure*}

\begin{figure*}
    \centering
    \includegraphics[width=0.98\linewidth]{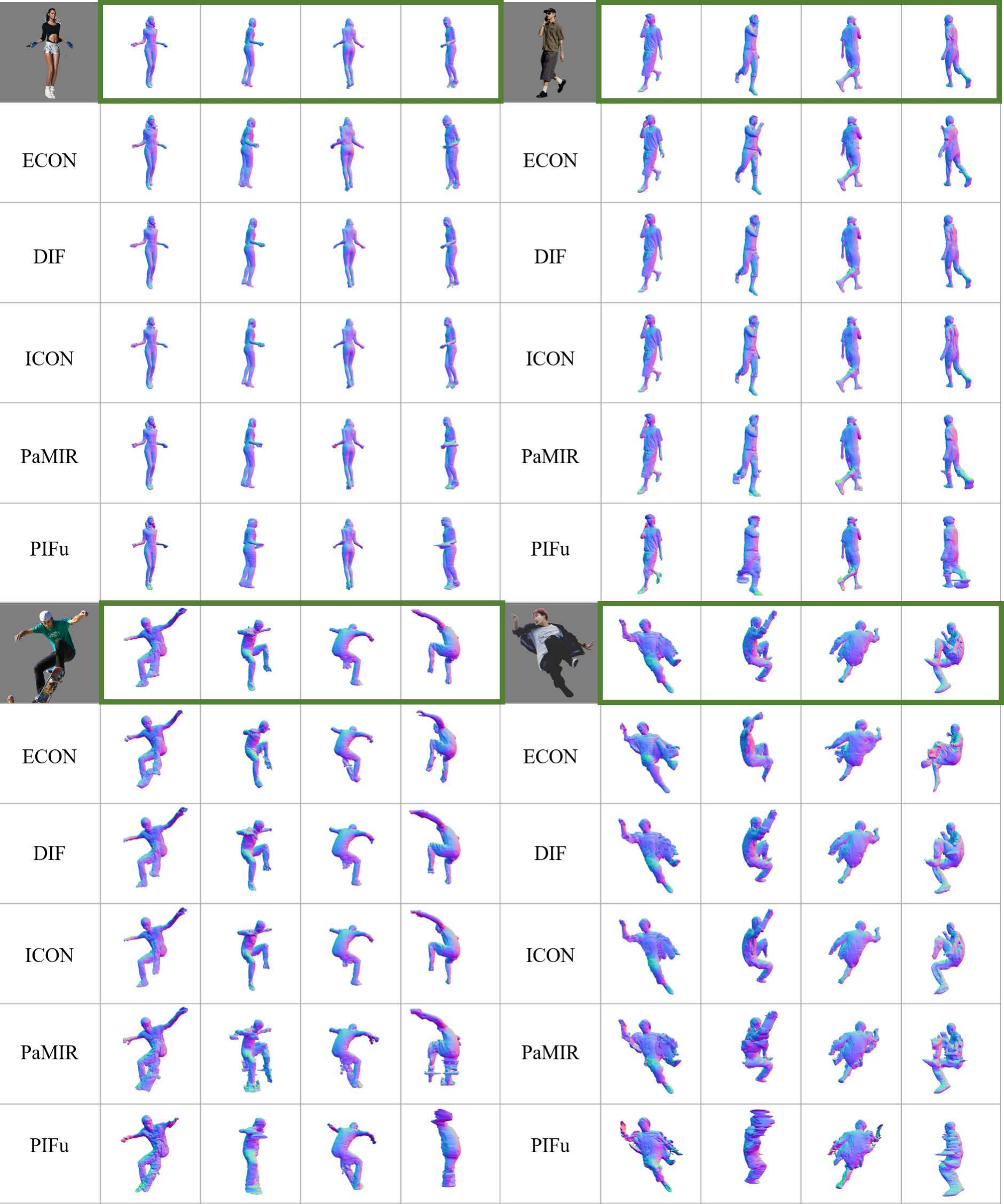}
    \caption{Visualization comparisons of reconstruction for our \textcolor[RGB]{84,130,53}{\sexyname~} vs SOTA.}
    \label{fig:sup_comp2_1}
\end{figure*}

\begin{figure*}
    \centering
    \includegraphics[width=0.98\linewidth]{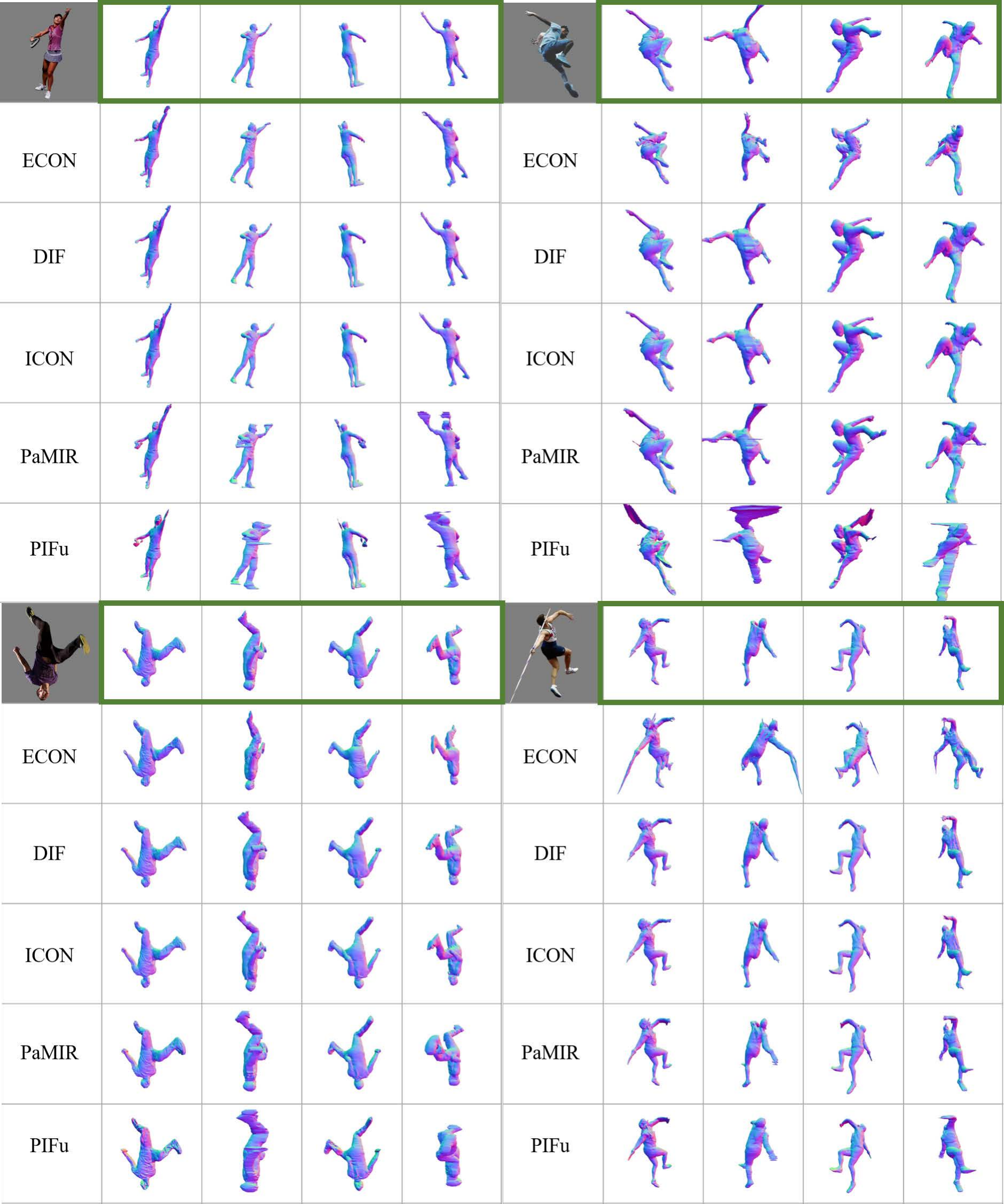}
    \caption{Visualization comparisons of reconstruction for our \textcolor[RGB]{84,130,53}{\sexyname~} vs SOTA.}
    \label{fig:sup_comp2_2}
\end{figure*}

\begin{figure*}
    \centering
    \includegraphics[width=0.98\linewidth]{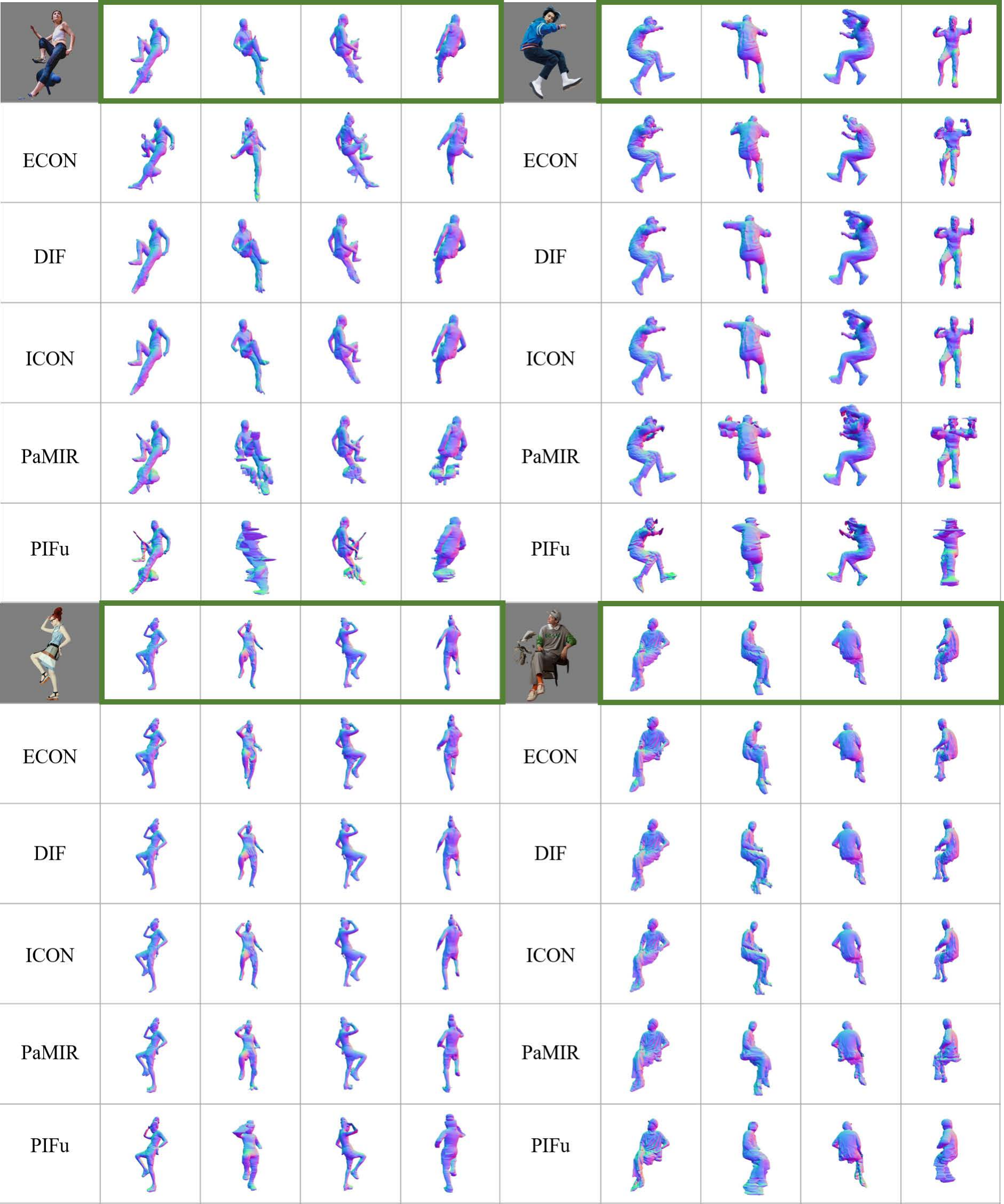}
    \caption{Visualization comparisons of reconstruction for our \textcolor[RGB]{84,130,53}{\sexyname~} vs SOTA.}
    \label{fig:sup_comp2_3}
\end{figure*}

\begin{figure*}
    \centering
    \includegraphics[width=0.98\linewidth]{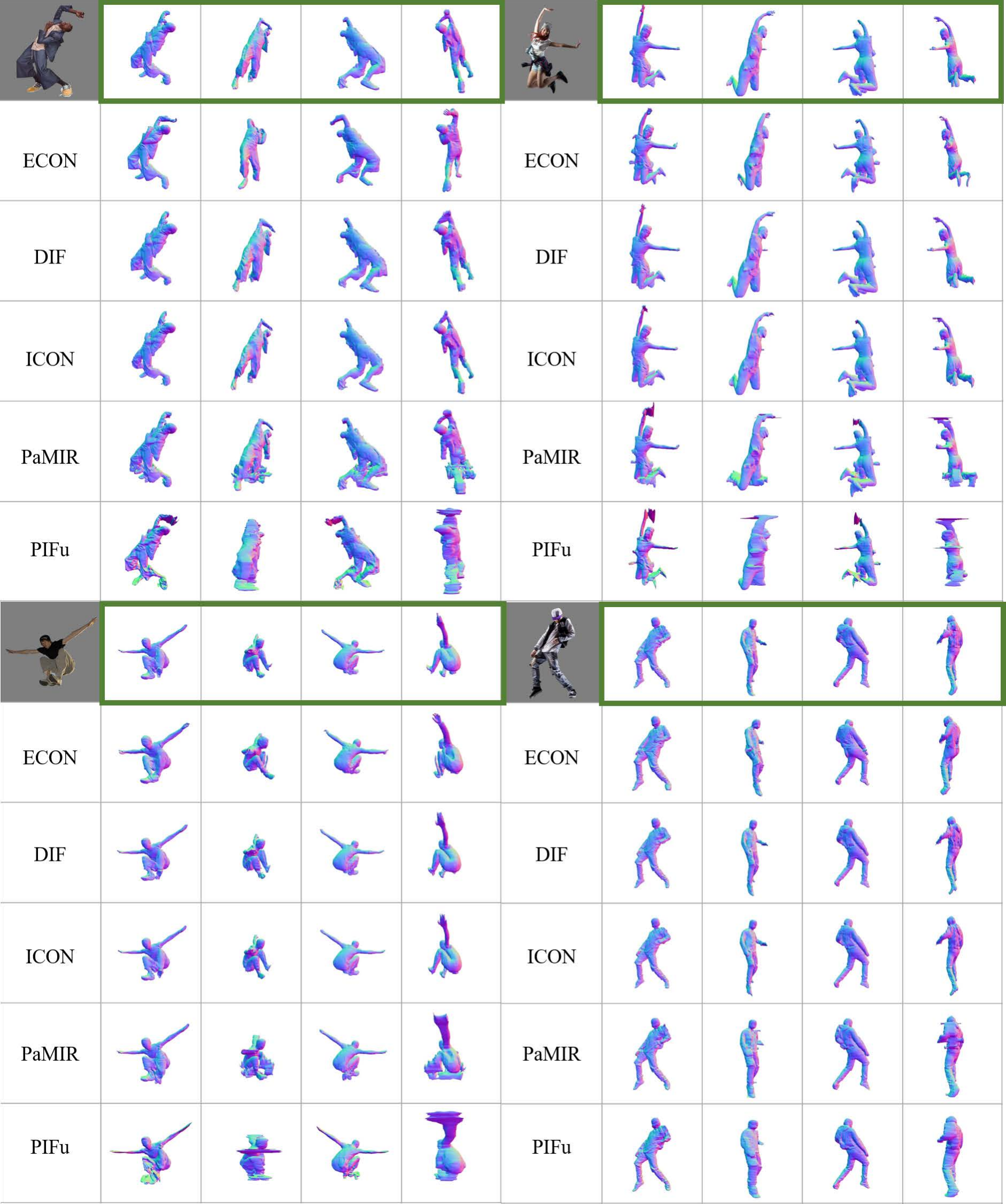}
    \caption{Visualization comparisons of reconstruction for our \textcolor[RGB]{84,130,53}{\sexyname~} vs SOTA.}
    \label{fig:sup_comp2_4}
\end{figure*}

We report more comparisons with state-of-the-art methods on in-the-wild images in 
Fig.~\ref{fig:sup_comp1_1}, Fig.~\ref{fig:sup_comp1_2}, Fig.~\ref{fig:sup_comp1_3}, Fig.~\ref{fig:sup_comp1_4},
Fig.~\ref{fig:sup_comp1_5},
Fig.~\ref{fig:sup_comp2_1},
Fig.~\ref{fig:sup_comp2_2},
Fig.~\ref{fig:sup_comp2_3},
Fig.~\ref{fig:sup_comp2_4}.
We render the reconstructed 3D models from four different views, \ie~$0^{\circ}, 90^{\circ}, 180^{\circ}, 270^{\circ}$.

\clearpage
{
    \small
}

\end{document}